\documentclass{article} %
\usepackage{colm2024_conference}

\usepackage{booktabs}
\usepackage{graphicx}
\usepackage{enumitem}
\usepackage{wrapfig}
\usepackage{algorithm}
\usepackage{algpseudocode}

\usepackage{microtype}
\usepackage{amsmath}
\usepackage{colortbl}
\usepackage[utf8]{inputenc}
\definecolor{lightgray}{rgb}{0.9,0.9,0.9}
\usepackage{caption}
\usepackage{subcaption}
\usepackage{xcolor}
\usepackage{setspace}
\usepackage{url}
\usepackage{multirow}
\usepackage{colortbl}
\usepackage{tabularx}
\usepackage{blindtext}
\usepackage{pgfplots}
\pgfplotsset{compat=1.18} 
\usepackage{tikz}
\usetikzlibrary{er,positioning,bayesnet}
\usepackage{makecell}
\usepackage{siunitx}
\usepackage{nicefrac}
\usepackage{tocloft}
\usepackage{listings}
\usepackage[raster,skins]{tcolorbox} %
\usepackage{xltabular}
\usepackage{adjustbox}
\usepackage{xurl}
\usepackage{arydshln}

\definecolor{lightgray}{rgb}{0.9,0.9,0.9}
\definecolor{safetygroup}{HTML}{B9A8C2}
\definecolor{generalgroup}{HTML}{E3D8D9}
\definecolor{reasongroup}{HTML}{B0BBCB}
\definecolor{featuresynthbase}{HTML}{EB687B}   
\definecolor{randomsynthbase}{HTML}{F1837B}    
\colorlet{featuresynthrow}{featuresynthbase!30}
\colorlet{randomsynthrow}{randomsynthbase!15}

\title{Qwen-Scope: Turning Sparse Features into\\Development Tools for Large Language Models}

\author{
\bf Qwen Team
}

\begin{document}

\maketitle

\begin{abstract}

Large language models have achieved remarkable capabilities across diverse tasks, yet their internal decision-making processes remain largely opaque, limiting our ability to inspect, control, and systematically improve them. This opacity motivates a growing body of research in mechanistic interpretability, with sparse autoencoders (SAEs) emerging as one of the most promising tools for decomposing model activations into sparse, interpretable feature representations.
We introduce \textbf{Qwen-Scope}, an open-source suite of SAEs built on the Qwen model family, comprising 14 groups of SAEs across 7 model variants from the Qwen3 and Qwen3.5 series, covering both dense and mixture-of-expert architectures.
Built on top of these SAEs, we show that SAEs can go beyond post-hoc analysis to serve as practical interfaces for model development along four directions: (i) \textbf{inference-time steering}, where SAE feature directions control language, concepts, and preferences without modifying model weights; (ii) \textbf{evaluation analysis}, where activated SAE features provide a representation-level proxy for benchmark redundancy and capability coverage; (iii) \textbf{data-centric workflows}, where SAE features support multilingual toxicity classification and safety-oriented data synthesis; and (iv) \textbf{post-training optimization}, where SAE-derived signals are incorporated into supervised fine-tuning and reinforcement learning objectives to mitigate undesirable behaviors such as code-switching and repetition. Together, these results demonstrate that SAEs can serve not only as post-hoc analysis tools, but also as reusable representation-level interfaces for diagnosing, controlling, evaluating, and improving large language models. 
By open-sourcing Qwen-Scope, we aim to support mechanistic research and accelerate practical workflows that connect model internals to downstream behavior.

\end{abstract}

\begin{figure}[h]
  \centering
  \includegraphics[width=0.9\linewidth]{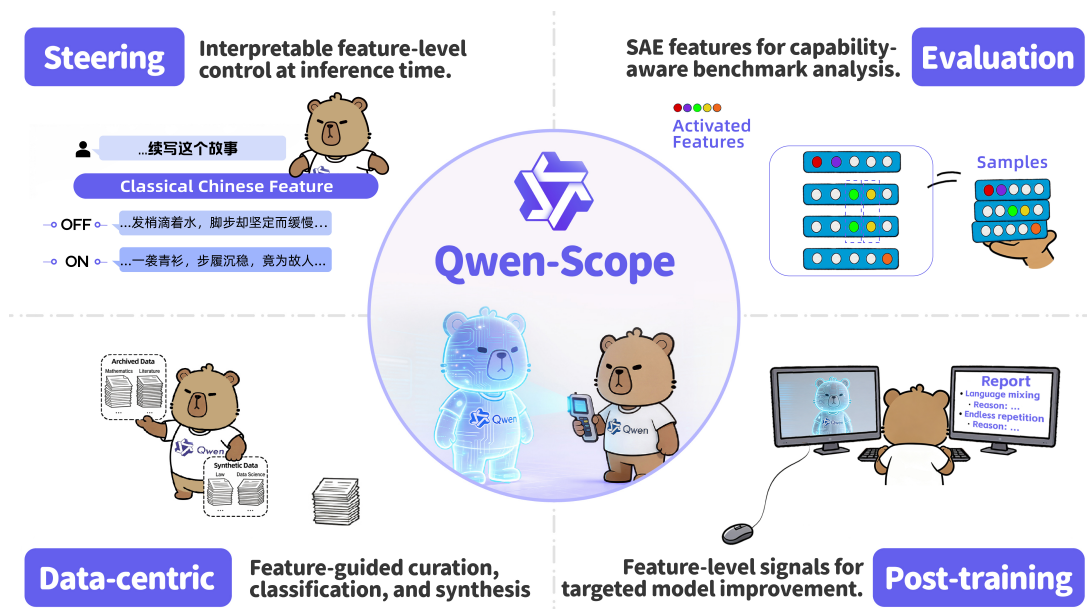}
  \caption{\textbf{Overview of Qwen-Scope.} SAEs trained on Qwen3/3.5 serve as a common interface for four practical directions: interpretable steering at inference time, capability-aware benchmark analysis, feature-guided data workflows, and targeted post-training for model improvement.}
  \label{fig:overview}
\end{figure}

\clearpage

\tableofcontents
\clearpage

\section{Introduction}

Large language models (LLMs) have achieved remarkable capabilities across a wide range of tasks, from natural language understanding and generation to complex reasoning, coding, and mathematical problem solving~\citep{deepseekr1,qwen3,singh2025openai,claude46,comanici2025gemini25pushingfrontier}. Despite their impressive performance, LLMs remain largely opaque systems whose internal decision-making processes are poorly understood, raising concerns about their reliability and trustworthiness~\citep{naseem2026mechanistic,shu-etal-2025-survey}. This opacity motivates a growing body of research in mechanistic interpretability, which aims to reverse-engineer the internal computations of LLMs~\citep{singh2024rethinkinginterpretabilityeralarge,jacob2024transformersfindniterpretablellmfeaturecircuits,bereska2024mechanistic,sharkey2025open}.

Sparse autoencoders (SAEs) have emerged as a promising tool for mechanistic interpretability in LLMs~\citep{cunningham2023sparse,gao2024scalingevaluatingsparseautoencoders,bricken2023monosemanticity}. Intuitively, an SAE learns a large dictionary of sparse latent features that reconstruct a model's internal activation vectors. Each input activates only a small subset of these features, making it possible to describe a high-dimensional hidden state in terms of a small number of more interpretable directions~\citep{elhage2022superposition,park2023linear,nanda-etal-2023-emergent}. SAEs address this by learning an overcomplete feature basis, so that an activation can be approximately reconstructed from a sparse set of learned feature directions. In this sense, SAE features provide a candidate vocabulary for describing what information is present in a model's internal state.

However, the prevailing SAE workflow still treats features primarily as objects of post-hoc analysis: researchers discover, inspect, and label features, but the connection from these features to concrete model-development workflows remains underexplored~\citep{shu-etal-2025-survey,sharkey2025open}. We argue that interpretability should move beyond description and become a practical interface for controlling, auditing, and improving LLMs.

In this work, we introduce \textbf{Qwen-Scope}, an open-source SAE suite built upon the Qwen family of models, together with a set of practical applications demonstrating how SAE features can be leveraged to control, audit, and improve language models. We release 14 groups of SAEs covering 7 model variants from the Qwen3 and Qwen3.5 series, encompassing both dense and mixture-of-experts (MoE) architectures. Built on top of these SAEs, we demonstrate four categories of applications:
\begin{enumerate}
    \item \textbf{Steering}: as the most widely adopted application of SAEs in prior work, we show that model behavior can be influenced through feature-level interventions, enabling control over language, concepts, and preferences without modifying model weights.
    \item \textbf{Evaluation}: we explore the use of feature coverage as a proxy for full-scale benchmarking and use it to study redundancy and representational concentration across evaluation sets.
    \item \textbf{Data-centric workflows}: we apply SAE features to multilingual toxicity classification and safety-oriented data synthesis.
    \item \textbf{Post-training}: we leverage SAE features to guide both SFT and RL. In SFT, we suppress language-specific feature activations via an auxiliary loss to reduce code-switching. In RL, we steer repetition-related features to synthesize rare negative rollouts, providing explicit training signals against endless repetition.
\end{enumerate}

Collectively, these results show that SAEs are not only tools for post-hoc inspection, but can also serve as a reusable representation-level interface for model development. Through Qwen-Scope, the same set of interpretable features can be used to diagnose model behavior, steer outputs, analyze evaluation data, guide data construction, and improve post-training.

The remainder of this paper is organized as follows. Section~\ref{sec:training} describes the construction of Qwen-Scope, including model coverage, SAE training procedures, and implementation details. Section~\ref{sec:steering} presents empirical studies on inference-time steering with SAE features. Section~\ref{sec:eval} analyzes redundancy and capability overlap within and across evaluation benchmarks using feature coverage. Sections~\ref{sec:applications_classification} and~\ref{sec:applications_synthesis} present two data-centric applications: data classification and data synthesis. Sections~\ref{sec:sft} and~\ref{sec:rl} show how Qwen-Scope can be used in post-training, including supervised fine-tuning and reinforcement learning. Section~\ref{sec:conclusion} concludes with the main contributions of Qwen-Scope and discusses its broader impact.

Finally, Qwen-Scope is intended as an open foundation for community-driven interpretability research on the Qwen model family. By releasing these SAE modules and demonstrating their practical use across steering, evaluation, data workflows, and post-training, we hope to enable researchers and developers to explore Qwen-series models more deeply, uncover new internal mechanisms, and discover additional valuable applications beyond those presented in this report.
\clearpage

\section{Training in Practice}
\label{sec:training}

\begin{table}[h]
    \centering
    \small
    \setlength{\tabcolsep}{4pt}
        \caption{Overview of all released sparse autoencoders (SAEs) in Qwen-Scope. In total, we release 14 groups of SAE weights across 7 Qwen backbones, covering both dense and mixture-of-experts (MoE) architectures. For each backbone, SAEs are trained on all layers. Unless otherwise specified, the SAEs are trained on the corresponding base model; Qwen3.5-27B is the only backbone whose SAEs are trained on the instruct variant. For MoE models, we additionally release wider SAEs to capture more fine-grained features. Expansion factor: the ratio of SAE width to the hidden size.}
    \resizebox{\linewidth}{!}{%
    \begin{tabular}{llcccccc}
    \toprule
    Architecture & Model & Backbone type & Trained layers & Hidden size & SAE width & Expansion factor & Top-$k$ ($L_0$) \\
    \midrule
    \multirow{5}{*}{Dense}
        & SAE-Res-Qwen3-1.7B-Base-W32K-L0\_\{50,100\}  & Base     & 1--28 (all) & 2048 & 32K & 16 & \{50, 100\} \\
        & SAE-Res-Qwen3-8B-Base-W64K-L0\_\{50,100\}    & Base     & 1--36 (all) & 4096 & 64K & 16 & \{50, 100\} \\
        \cdashline{2-8}[1pt/2.5pt]\noalign{\vskip 0.5ex}
        & SAE-Res-Qwen3.5-2B-Base-W32K-L0\_\{50,100\}  & Base     & 1--24 (all) & 2048 & 32K & 16 & \{50, 100\} \\
        & SAE-Res-Qwen3.5-9B-Base-W64K-L0\_\{50,100\}  & Base     & 1--32 (all) & 4096 & 64K & 16 & \{50, 100\} \\
        & SAE-Res-Qwen3.5-27B-W80K-L0\_\{50,100\} & Instruct & 1--64 (all) & 5120 & 80K & 16 & \{50, 100\} \\
    \midrule
    \multirow{4}{*}{MoE}
        & SAE-Res-Qwen3-30B-A3B-Base-W32K-L0\_50
            & \multirow{2}{*}{Base}
            & \multirow{2}{*}{1--48 (all)}
            & \multirow{2}{*}{2048}
            & 32K & 16 & 50 \\
        & SAE-Res-Qwen3-30B-A3B-Base-W128K-L0\_100 & & & & 128K & 64 & 100 \\
        \cdashline{2-8}[1pt/2.5pt]\noalign{\vskip 0.5ex}
        & SAE-Res-Qwen3.5-35B-A3B-Base-W32K-L0\_50
            & \multirow{2}{*}{Base}
            & \multirow{2}{*}{1--40 (all)}
            & \multirow{2}{*}{2048}
            & 32K & 16 & 50 \\
        & SAE-Res-Qwen3.5-35B-A3B-Base-W128K-L0\_100 & & & & 128K & 64 & 100 \\
    \bottomrule
    \end{tabular}%
    }

    \label{tab:train_stats}
\end{table}

\subsection{Why Sparse Auto-Encoders?}
Sparse Autoencoders (SAEs) have emerged as a foundational tool for learning disentangled, interpretable representations in high-dimensional neural activations~\citep{lieberum-etal-2024-gemma,he2024llamascopeextractingmillions}. Unlike conventional autoencoders that prioritize reconstruction fidelity alone, SAEs explicitly enforce sparsity in the latent space, encouraging each latent dimension to activate only for a narrow subset of inputs. Beyond interpretability, this sparse structure has made SAEs increasingly useful as a practical interface for model intervention and analysis, with recent work applying them to steering~\citep{steer1,steer2}, targeted unlearning~\citep{unlearn1,unlearn2}, and reasoning-related representations~\citep{reason1,reason2,fang2026controllable}. Motivated by these applications, we build a corresponding SAE toolkit for the Qwen family to support both mechanistic analysis and practical downstream use. 

\subsection{Training in Practice}
We train SAEs for the Qwen3 and Qwen3.5 model families. Our release provides layer-wise sparse representations for both dense and mixture-of-experts (MoE) backbones under a unified training pipeline. For each backbone and transformer layer, we collect residual-stream activations and train a separate SAE to reconstruct these activations with a sparse set of latent features. Thus, each released SAE provides a feature basis for a specific layer of a specific model, enabling downstream analysis and intervention at the level of SAE feature activations rather than raw hidden states. Table~\ref{tab:train_stats} summarizes the full release scope, including the backbone type, trained layers, hidden size, SAE width, expansion factor, and sparsity level used for each model.

As shown in Table~\ref{tab:train_stats}, our release covers all transformer layers of 7 Qwen backbones and includes 14 groups of SAE weights in total. We train all SAEs sampled from in-house pretraining data. During training, the SAE encoder maps each residual-stream activation to an overcomplete latent representation, and a Top-\(k\) activation rule keeps only the largest \(k\) latent activations for reconstruction. We release SAEs with Top-\(k\) values of 50 or 100. For dense backbones, the SAE width scales with the model hidden size; for MoE backbones, we additionally release wider SAEs, up to \(64\times\) the hidden size, to capture more fine-grained representation structure.

To maintain training stability, we apply the following settings:
\begin{itemize}
    \item We apply an auxiliary loss with weight \(\frac{1}{32}\), following~\cite{gao2024scalingevaluatingsparseautoencoders}, to reduce the fraction of dead features. By the end of training, almost all released SAEs have a negligible number of dead features.
    
    \item We filter out activations with extremely large \(L_2\)-norm values, following~\cite{marks2024dictionary_learning}, to stabilize the reconstruction objective. These outliers appear most often for Qwen3-1.7B and Qwen3-8B, especially in activations associated with the first token of each input sequence.
\end{itemize}

This training setup yields a collection of layer-wise SAE feature dictionaries that are reused throughout the report for steering, evaluation analysis, data-centric workflows, and post-training applications.

\clearpage

\section{Application: Steering with SAEs during Inference}
\label{sec:steering}

\begin{figure}[h]
    \centering
    \includegraphics[width=0.9\linewidth]{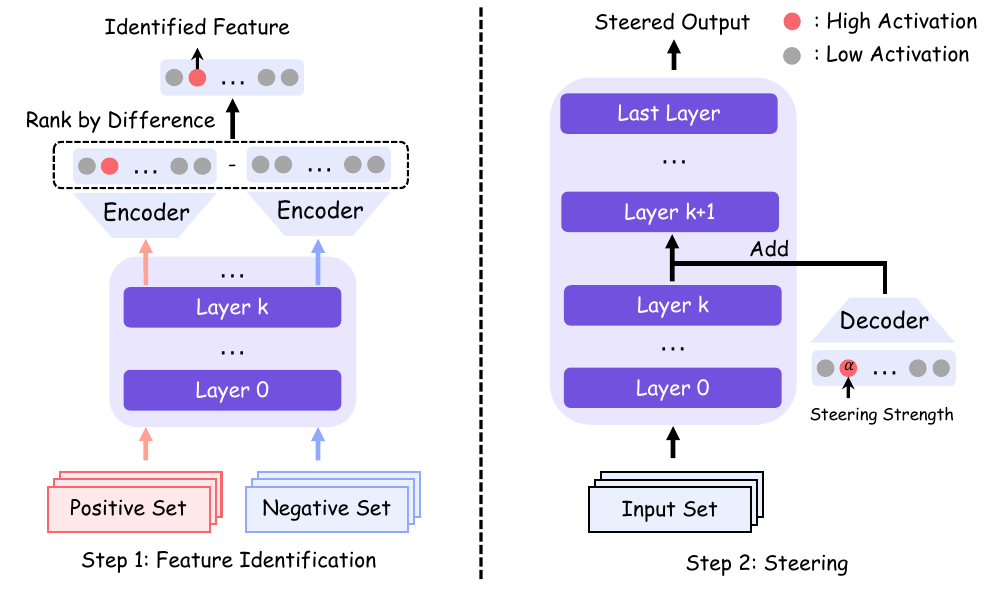}
    \caption{Illustration of the two-step SAE-based steering pipeline: (1) contrastive feature identification, where SAE activations are compared between positive and negative example sets to identify the most discriminative feature directions; and (2) steering, where the identified feature is injected into the model's hidden state via Equation~\ref{eq:steering_ill}.}
    \label{fig:steering_illustration}
\end{figure}

\subsection{What is Steering?}

Steering is based on the hypothesis that high-level concepts, skills, or behaviors are encoded as directions in the model's internal representation space. Under this view, intervening on a hidden state along a specific direction can move the model's internal computation toward the corresponding concept, thereby influencing the final output without updating model parameters~\citep{zhang2026locate,rimsky-etal-2024-steering}. 

SAEs are especially well-suited for this purpose because they decompose model activations into sparse and more interpretable features, making it possible to associate individual directions with more specific behaviors or semantic properties. Once a feature of interest is identified, we can steer the model by adding or suppressing the corresponding feature direction in the residual stream. A common form of feature steering can be written as:
\begin{equation}
    \mathbf{h}' \leftarrow \mathbf{h} + \alpha \mathbf{d},
    \label{eq:steering_ill}
\end{equation}
where \(\mathbf{h}\) is the original hidden state of the model, \(\mathbf{d}\) is the SAE feature direction, and \(\alpha\) controls the strength of the intervention. Positive values of \(\alpha\) amplify the feature, while negative values suppress it. After replacing \(\mathbf{h}\) with \(\mathbf{h}'\), the model continues the forward pass with the modified representation, which can lead to changes in the generated output.

\subsection{How to Identify Features for Steering}

Existing methods for finding SAE features to steer can be roughly grouped into two types: \textbf{contrastive methods} and \textbf{automatic interpretation methods}.

Contrastive methods begin by defining a target concept or behavior of interest, such as a language, a style, or a preference. The next step is to construct two groups of examples: a positive set that strongly exhibits the target property, and a negative or neutral set that does not. The activations from these examples are then passed through the SAE encoder to obtain feature activations. By comparing the average activation of each feature across the two groups, one can identify features that are selectively associated with the target property. Features with the largest activation differences are then treated as the most relevant candidates for steering~\citep{he2025saif,bayat2025steering,deng2025unveilinglanguagespecificfeatureslarge,shi-etal-2025-route}.

Automatic interpretation methods take a more direct approach by trying to assign human-readable meanings to SAE features. Instead of first defining a target behavior and searching for discriminative features, these methods start from the features themselves. For each feature, one collects the text contexts in which it activates strongly, and then provides these activating examples to a stronger language model. The language model is prompted to summarize the shared pattern across these examples and produce a short natural-language description of what the feature appears to represent~\citep{DBLP:conf/icml/PauloMJB25}. This makes it possible to interpret and organize very large numbers of SAE features at scale, and the resulting descriptions can help researchers quickly identify features that are relevant for downstream steering.

\subsection{Case Studies of SAE Steering}

\begin{figure}[t]
    \centering
    \includegraphics[width=1.0\linewidth]{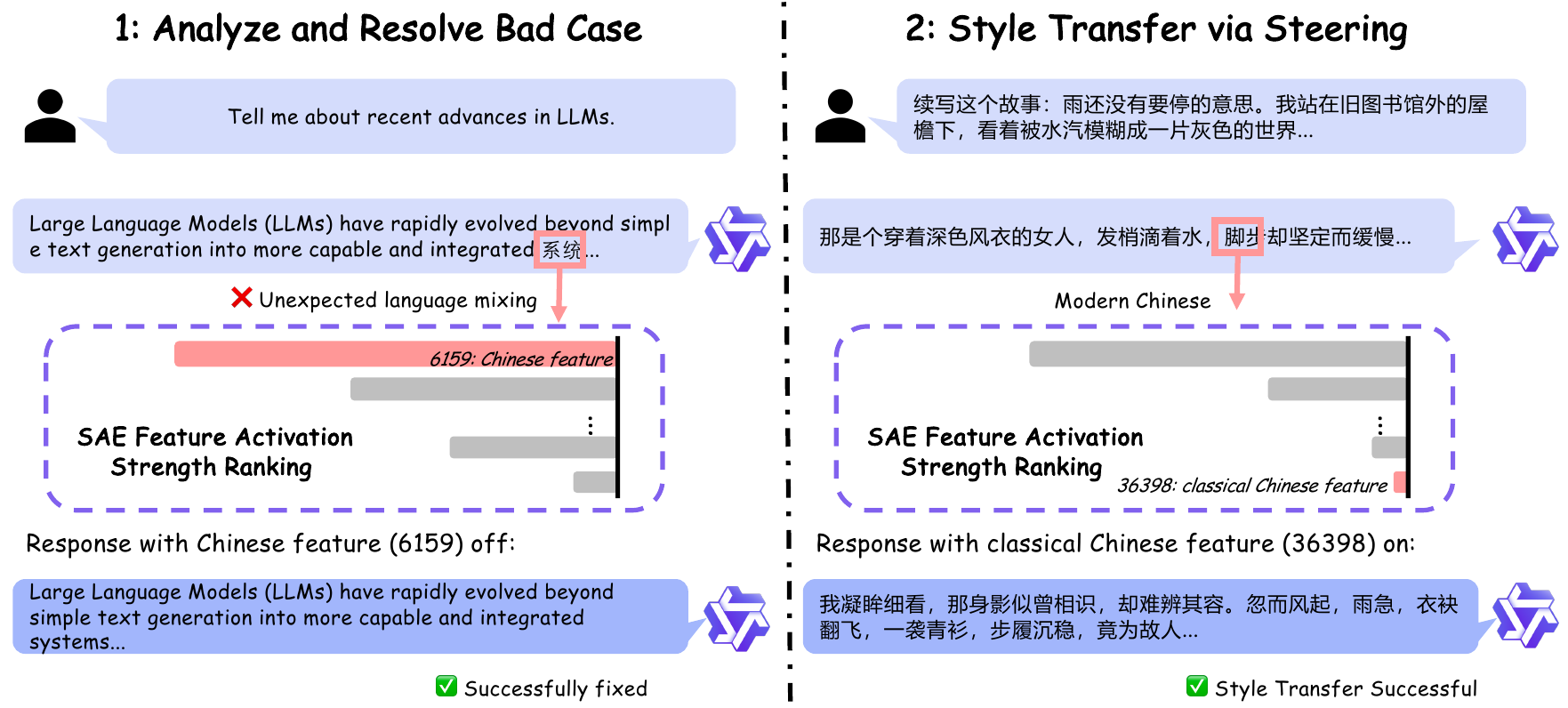}
    \caption{
    \textbf{SAE features provide interpretable handles for model analysis and control.}
    \textbf{Left:} SAE activations can be used to undesirable generation behavior. When the model is prompted in English, the response unexpectedly mixes in Chinese text. Ranking SAE features by activation strength reveals a highly activated Chinese-language feature (id: 6159). Suppressing this feature during generation removes the unexpected language mixing while preserving the intended English response.
    \textbf{Right:} The same feature-level interface can also be used for controlled style transfer. Given a modern Chinese continuation task, activating a classical-Chinese feature (id: 36398) steers the model toward a classical literary style.}
    \label{fig:sae_feature_debugging_and_steering}
\end{figure}

To illustrate how SAE-based steering works in practice, we present two representative case studies using Qwen3 models, as shown in Figure~\ref{fig:sae_feature_debugging_and_steering}. These examples highlight two complementary uses of SAE features: diagnosing undesirable behavior by identifying the responsible internal feature, and controlling generation by activating a desired feature direction.

\paragraph{Analyzing and Resolving Bad Cases.}
In the first example, the model is prompted in English but unexpectedly mixes in Chinese text during generation. By ranking SAE features according to their activation strength on the problematic response, we identify a highly activated Chinese-language feature. This provides an interpretable explanation of the failure: the model has entered an internal direction associated with Chinese generation. Suppressing this feature during inference removes the unexpected language mixing and restores the intended English response. This demonstrates that SAE features can serve as diagnostic handles for tracing and correcting undesirable generation behavior.

\paragraph{Style Transfer via Steering.}
In the second example, the model is asked to continue a story written in modern Chinese. By activating an SAE feature associated with classical Chinese, the model shifts its continuation toward a classical literary style while preserving the semantic direction of the prompt. This shows that SAE features can also be used constructively: instead of only suppressing unwanted behavior, they can steer generation toward a desired style or linguistic register.

Together, these examples show that SAE steering provides an interpretable mechanism for both model debugging and controllable generation. Because the intervention operates directly on feature directions in the residual stream, it can modify generation behavior without updating model weights.

\clearpage

\section{Application: Evaluation}
\label{sec:eval}
\begin{figure}[h]
    \centering
    \includegraphics[width=0.95\linewidth]{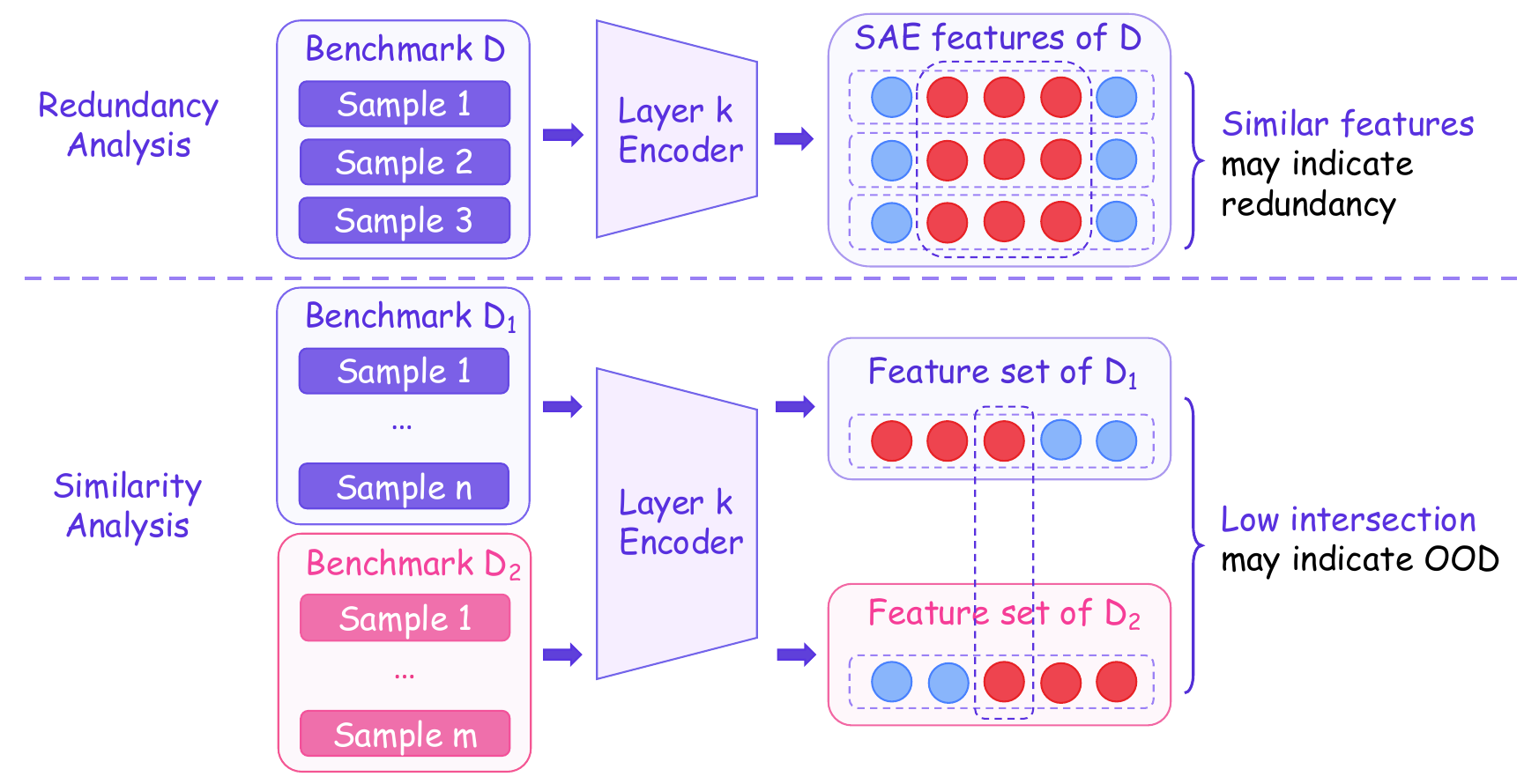}
    \caption{Illustration of the proposed SAE-based benchmark analysis framework, covering feature extraction, intra-benchmark redundancy measurement, and inter-benchmark similarity analysis.}
    \label{fig:eval_pipeline}
\end{figure}
The rapid expansion of LLM evaluation benchmarks raises two practical questions:
(1)~given a benchmark with $N$ samples, can a small subset $\mathcal{S} \subset \mathcal{D}$ of size $n \ll N$ preserve the model ranking induced by the full dataset;
(2)~given two benchmarks, do they probe the same capabilities or genuinely different ones, and can we answer this \emph{without} running any model evaluation?

The direct approach --- evaluating a panel of $M$ models on every benchmark and subset --- requires $\mathcal{O}(M \times N)$ forward passes and is prohibitively expensive for large-scale benchmark curation.
We observe that Sparse Autoencoders provide a natural alternative.
When a model processes a benchmark sample, the SAE decomposes the resulting activation into a sparse set of active features, each interpretable as a ``micro-capability.''
The set of features activated by a benchmark thus constitutes a compact fingerprint of what it probes.
A benchmark is \emph{redundant} if many samples activate the same features (coverage saturates early); two benchmarks are \emph{similar} if they activate largely overlapping feature sets.

Building on this intuition, we propose a unified framework for benchmark curation that leverages SAE-derived feature representations as a proxy for model-level evaluation. We first introduce the SAE-based feature extraction framework (Section~\ref{sec:eval_sae}), then develop SAE feature-based redundancy metrics for single benchmarks (Section~\ref{sec:eval_redundancy}), and finally extend the framework to inter-benchmark similarity and out-of-distribution detection (Section~\ref{sec:eval_ood}). A schematic diagram of the pipeline is shown in Figure~\ref{fig:eval_pipeline}.

\subsection{SAE Feature Extraction}
\label{sec:eval_sae}
A benchmark $\mathcal{D} = \{x_1, x_2, \ldots, x_N\}$ is a collection of $N$ evaluation samples. For a given language model $\mathcal{M}$ equipped with an SAE at a chosen layer, we define the active feature set of sample $x_i$ as:
\begin{equation}
    F(x_i) = \bigl\{j \in \{1, \ldots, D\} : z_j(x_i) > 0\bigr\},
\end{equation}
where $z_j(x_i)$ is the $j$-th component of the SAE latent representation of $x_i$, extracted at the last token position. Note that $z_j(x_i)$ implicitly incorporates the \text{Top-k ReLU} activation applied within the SAE encoder; we omit this detail from the notation for brevity.
The feature footprint of the entire benchmark is:
\begin{equation}
    F(\mathcal{D}) = \bigcup_{i=1}^{N} F(x_i).
\end{equation}

\subsection{Benchmark Redundancy}
\label{sec:eval_redundancy}

\paragraph{Performance-based redundancy.}
The most direct way to measure redundancy is to ask: how small can a subset be while still preserving the model ranking? To illustrate this intuitively, consider the following two simple mathematical problems, drawn from GSM8K~\citep{gsm8k} and MATH~\citep{math}, respectively:
\begin{itemize}
    \item Candy has 15 light blue spools of thread, 45 dark blue
spools of thread, 40 light green spools of thread, and 50 dark green spools of thread. What percent of her spools are blue?
    \item Gina has five pairs of white socks, three pairs of black socks, and two pairs of red socks. What percent of her socks are red?
\end{itemize}
Both problems share an identical mathematical structure, which involves computing a ratio and expressing it as a percentage, and they differ only in surface context. As training corpora scale up, models become increasingly robust to surface-level context variation, rendering repeated evaluation on structurally identical problems redundant. For the purpose of model ranking, such samples contribute little discriminative power. To quantify the discriminative power of benchmark samples, we introduce the following framework. Fix a panel of $M$ models.
Let $p \in \mathbb{R}^M$ denote the vector of model accuracies on the full benchmark $\mathcal{D}$, and $\hat{p}(\mathcal{S})$ the corresponding vector on a subset $\mathcal{S}$.
We measure ranking agreement via Kendall's $\tau$:
\begin{equation}
    \tau(\mathcal{S}, \mathcal{D}) = \tau\!\bigl(p,\; \hat{p}(\mathcal{S})\bigr),
\end{equation}

Kendall's $\tau$ is preferred over Spearman's $\rho$ here because it has a direct combinatorial interpretation: $(\tau + 1)/2$ equals the fraction of model pairs whose relative ordering is preserved by the subset. For a single random subset, $\tau(\mathcal{S}, \mathcal{D})$ is a random variable.
To characterize the typical behavior at each subset size, we take expectations:
\begin{equation}
    \tau_n = \mathbb{E}_{\mathcal{S} \subseteq \mathcal{D},\, |\mathcal{S}|=n}\!\bigl[\tau(\mathcal{S}, \mathcal{D})\bigr].
\end{equation}
The curve $n \mapsto \tau_n$ is the benchmark's \emph{redundancy profile}: it starts near zero for very small $n$ and approaches 1 as $n \to N$.
A curve that saturates early indicates that most samples are interchangeable for ranking purposes.
To obtain a single scalar summary, we take the area under this curve:
\begin{equation}
    \mathcal{R}(\mathcal{D}) = \frac{1}{N} \sum_{n=1}^{N} \tau_n.
\end{equation}

A higher $\mathcal{R}$ means the benchmark is more redundant; in other words, fewer samples suffice to recover the full ranking.

\paragraph{Limitation of performance-based redundancy.}
Computing $\mathcal{R}(\mathcal{D})$ requires evaluating all $M$ models on the full benchmark: obtaining $\tau_n$ at even a single value of $n$ demands sampling many random subsets and running model evaluations on each.
This is precisely the cost we set out to avoid.
We therefore ask: \emph{can we estimate benchmark redundancy without any model evaluation?}

\paragraph{SAE feature-based redundancy.}
We propose a feature-based proxy that depends only on the SAE feature structure, requiring no model evaluation.
The key idea is to replace the rank-correlation curve $n \mapsto \tau_n$ with a feature-coverage curve: as we add samples to a random subset, how quickly does the set of activated features saturate?
Concretely, we define the expected feature coverage at size $n$ as:
\begin{equation}
    c_n = \mathbb{E}_{\mathcal{S} \subseteq \mathcal{D},\, |\mathcal{S}|=n}\!\left[\frac{|F(\mathcal{S})|}{|F(\mathcal{D})|}\right].
\end{equation}
The curve $n \mapsto c_n$ plays the same role as $n \mapsto \tau_n$: if a benchmark's feature coverage saturates quickly as we add samples, then its samples are redundant in the capability space (they activate largely the same features).
Aggregating via area under the curve gives a scalar analogue of $\mathcal{R}$:
\begin{equation}
    \mathrm{AUC}(c_n) = \frac{1}{N} \sum_{n=1}^{N} c_n.
\end{equation}

However, the raw coverage AUC alone does not capture absolute feature diversity. Consider two benchmarks of the same size $N$ whose coverage both grow linearly ($c_n = n/N$), yielding $\mathrm{AUC} = 0.5$ in both cases.
Suppose the first activates $|F(\mathcal{D})| = 1{,}000$ distinct features in total while the second activates $2{,}000$.
Both have the same AUC, yet the second benchmark clearly probes a broader range of capabilities; it should be considered less redundant.
The coverage curve, being normalized to $[0, 1]$, erases this difference in absolute scale. To restore it, we multiply the AUC by a growth-rate correction $N / |F(\mathcal{D})|$.
Intuitively, $|F(\mathcal{D})| / N$ measures how many new features each sample contributes on average: a benchmark that activates $2{,}000$ features over $N$ samples has twice the per-sample growth rate of one that activates $1{,}000$, and should therefore receive a lower redundancy score.
Multiplying by the reciprocal $N / |F(\mathcal{D})|$ achieves exactly this, yielding the \emph{feature redundancy}:
\begin{equation}
    \hat{\mathcal{R}}(\mathcal{D}) = \mathrm{AUC}(c_n) \cdot \frac{N}{|F(\mathcal{D})|} = \frac{\sum_{n=1}^{N} c_n}{|F(\mathcal{D})|}.
\end{equation}
This metric is high when two conditions hold simultaneously: (i)~feature coverage saturates quickly (high AUC), and (ii)~the feature growth rate is slow relative to the sample count (high $N / |F(\mathcal{D})|$).
Condition~(i) alone would unfairly favor small benchmarks; condition~(ii) alone would ignore the shape of the coverage curve. Their product balances both factors.

\begin{figure}[t]
    \centering
    \includegraphics[width=0.9\textwidth]{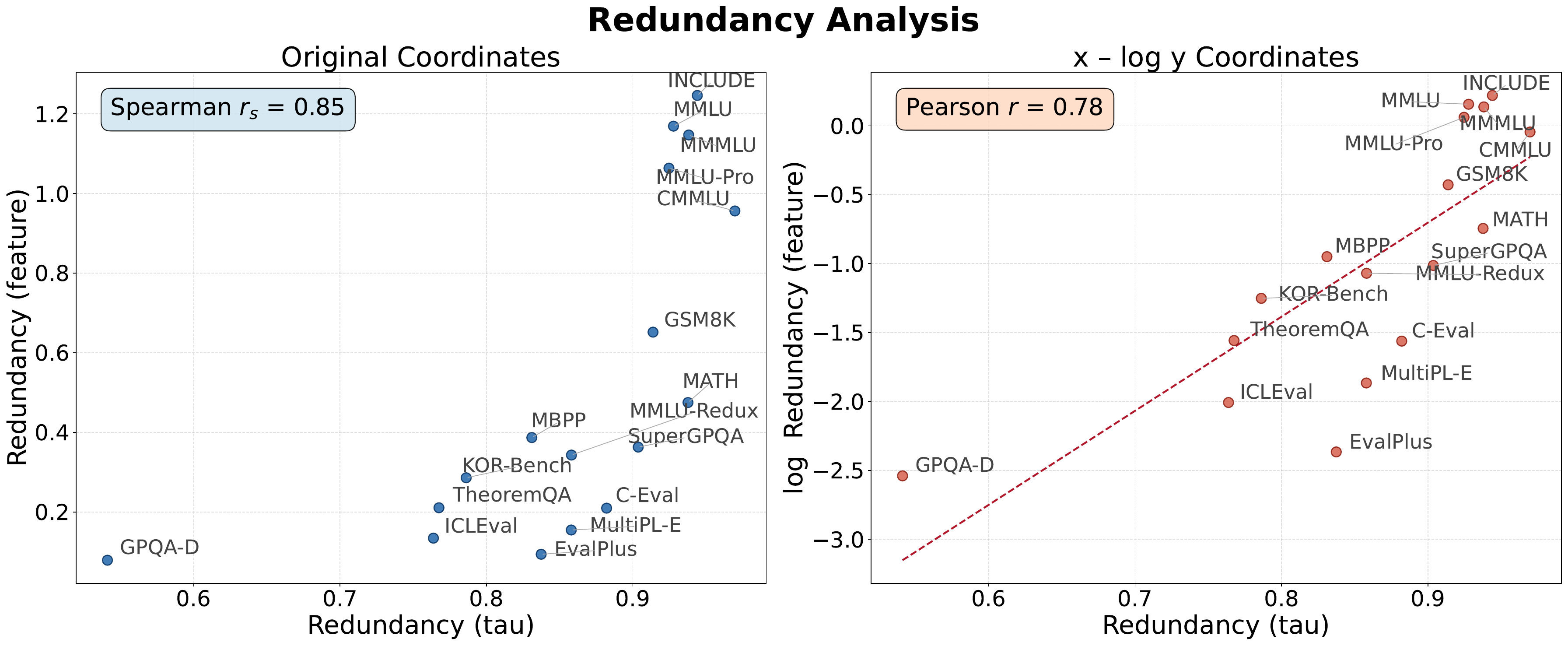}
    \caption{Spearman rank correlation between performance-based redundancy $\mathcal{R}(\mathcal{D})$ and feature redundancy $\hat{\mathcal{R}}(\mathcal{D})$ across 17 benchmarks (Spearman $\rho \approx 0.85$), suggesting that feature redundancy serves as a reasonable evaluation-free proxy for $\mathcal{R}(\mathcal{D})$.}
    \label{fig:redundancy_scatter}
\end{figure}

We select 26 pre-trained checkpoints with varying training steps and data mixture ratios, and evaluate the correlation between $\mathcal{R}(\mathcal{D})$ and $\hat{\mathcal{R}}(\mathcal{D})$ across 17 widely-used benchmarks spanning general knowledge, mathematics, coding, multilingual understanding, and in-context reasoning:
\begin{itemize}
    \item \textbf{General Tasks:} MMLU~\citep{mmlu}, MMLU-Redux~\citep{mmlu-redux}, MMLU-Pro~\citep{mmlu-pro}, SuperGPQA~\citep{supergpqa}, C-Eval~\citep{ceval}, CMMLU~\citep{cmmlu}.
    \item \textbf{STEM \& Math Tasks:} GSM8K~\citep{gsm8k}, MATH~\citep{math}, GPQA-Diamond~\citep{gpqa}, TheoremQA~\citep{theoremqa}.
    \item \textbf{Code Tasks:} MBPP~\citep{mbpp}, EvalPlus~\citep{evalplus}, MultiPL-E~\citep{multiple}.
    \item \textbf{Multilingual Tasks:} MMMLU~\citep{mmmlu}, INCLUDE~\citep{include}.
    \item \textbf{In-Context Reasoning Tasks:} KOR-Bench~\citep{korbench}, ICLEval~\citep{icleval}.
\end{itemize}
Key observations from the 17-benchmark analysis:
\begin{itemize}[leftmargin=2em, itemsep=2pt]
    \item The Spearman rank correlation between $\mathcal{R}(\mathcal{D})$ and $\hat{\mathcal{R}}(\mathcal{D})$ across 17 benchmarks is $\rho \approx 0.85$ (Figure~\ref{fig:redundancy_scatter}), suggesting that feature redundancy may serve as a reasonable evaluation-free proxy for performance-based redundancy.
    \item The correlation holds across benchmarks of vastly different sizes. For example, although GSM8K (1,319 samples) has fewer samples than MMLU-Redux (3,000 samples), it is positioned to the upper right of MMLU-Redux in the figure, indicating its inherent redundancy. Similarly, SuperGPQA contains 26,529 questions, yet exhibits relatively low redundancy.
\end{itemize}

These observations suggest that for benchmarks with high feature redundancy, only a small number of samples are needed to preserve the rankings of most models; for benchmarks with low feature redundancy, we may need to retain as many samples as possible, or even collect more evaluation data.

We note that high redundancy does not imply low benchmark quality.
Redundancy can be desirable: for example, to reduce evaluation variance or to ensure broad coverage within a specific domain.
The redundancy metric developed here is intended for a narrower operational scenario: when the goal is to rank models efficiently during iterative development, a highly redundant benchmark offers an opportunity to trade a modest amount of reliability for a significant reduction in evaluation cost.
Whether to exploit this trade-off is a decision that depends on the practitioner's priorities.

\subsection{Inter-Benchmark Similarity Analysis}
\label{sec:eval_ood}

We extend the framework to the inter-benchmark setting: given two benchmarks $\mathcal{D}_1$ and $\mathcal{D}_2$, do they probe the same capabilities?

\paragraph{Feature overlap.}
The feature footprint of a benchmark encodes what it probes; comparing two footprints therefore reveals whether two benchmarks test the same things.
We define the asymmetric feature overlap of $\mathcal{D}_1$ covered by $\mathcal{D}_2$ as:
\begin{equation}
    \mathrm{overlap}(\mathcal{D}_1, \mathcal{D}_2) = \frac{|F(\mathcal{D}_1) \cap F(\mathcal{D}_2)|}{|F(\mathcal{D}_1)|}.
\end{equation}

\begin{figure}[t]
    \centering
    \includegraphics[width=0.95\textwidth]{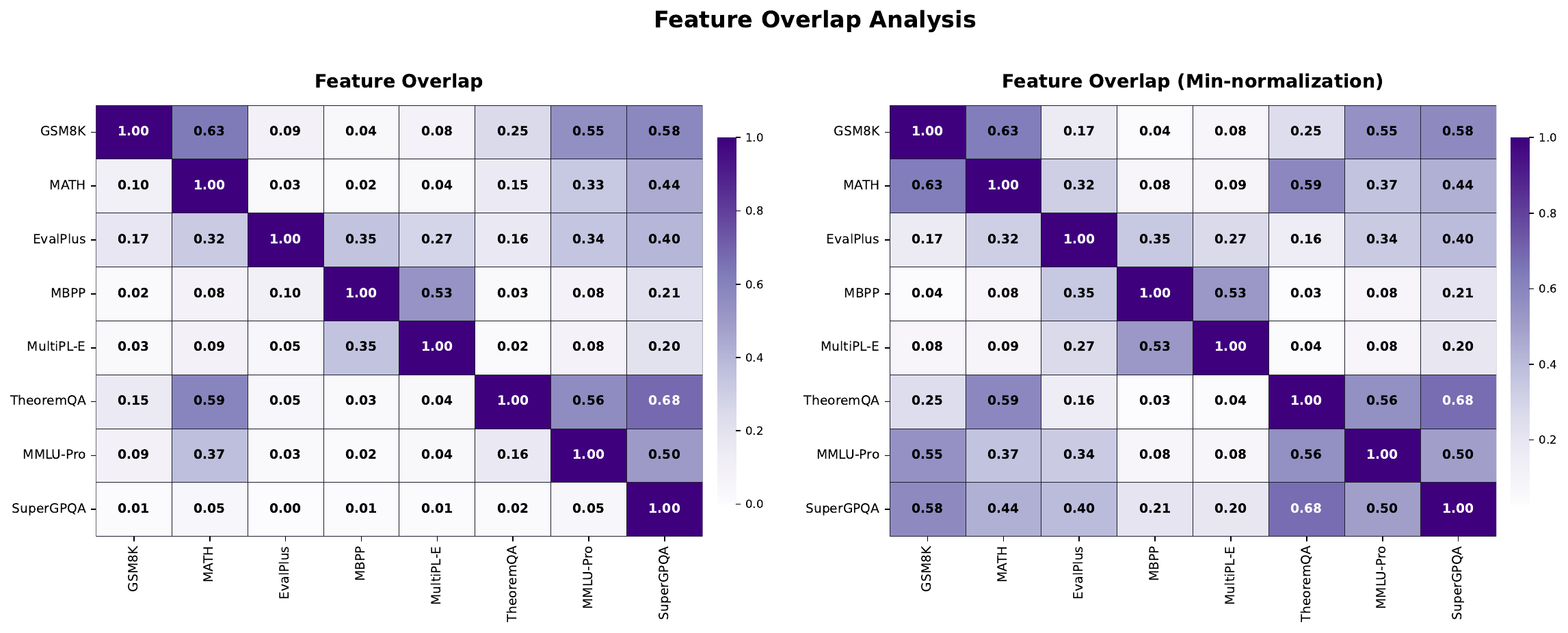} 
    \caption{Feature overlap matrix for eight benchmarks. Entry $(i,j)$ gives asymmetric overlap $\mathrm{overlap}(\mathcal{D}_i, \mathcal{D}_j)$ (left) and min-normalized overlap $\mathrm{overlap}_{\text{min}}(\mathcal{D}_i, \mathcal{D}_j)$ (right). The matrix reveals intuitive containment relationships: GSM8K is largely covered by MATH, code benchmarks form a tight cluster, and broad knowledge benchmarks subsume specialized ones.}
    \label{fig:overlap_heatmap}
\end{figure}

The asymmetry is deliberate and informative: it answers ``what fraction of $\mathcal{D}_1$'s capabilities are already covered by $\mathcal{D}_2$?'' For instance, $\mathrm{overlap}(\text{GSM8K}, \text{MATH})=0.63$ while $\mathrm{overlap}(\text{MATH}, \text{GSM8K}) = 0.10$ (Figure~\ref{fig:overlap_heatmap}), reflecting that elementary math capabilities are largely subsumed by competition math but not vice versa: MATH probes a much broader set of features that GSM8K does not touch.
The pairwise overlap matrix reveals intuitive structure: code benchmarks (EvalPlus, MBPP, MultiPL-E) form a cluster, and knowledge benchmarks (MMLU-Pro, SuperGPQA) subsume specialized ones like TheoremQA ($0.56-0.68$ coverage).

A natural question follows: Does this feature-level similarity translate into performance-level similarity?
That is, do benchmarks with high feature overlap also induce similar model rankings?

\paragraph{Symmetric overlap.}
To test this, we need symmetric metrics on both sides.
On the performance side, we use $\rho_\text{Pearson}(\mathcal{D}_1, \mathcal{D}_2) = \mathrm{corr}(p, q)$, the Pearson correlation between the two benchmarks' score vectors across models, which is naturally symmetric. A higher correlation characterizes the similarity between two benchmarks from a performance perspective.
On the feature side, we symmetrize via min-normalization:
\begin{equation}
    \mathrm{overlap}_{\min}(\mathcal{D}_1, \mathcal{D}_2) = \frac{|F(\mathcal{D}_1) \cap F(\mathcal{D}_2)|}{\min(|F(\mathcal{D}_1)|, |F(\mathcal{D}_2)|)}.
\end{equation}
The min-denominator ensures that the metric is high when the smaller benchmark's features are largely contained in the larger one, capturing the intuition of capability subsumption.

\paragraph{Direct correlation and its limitations.}
The direct correlation between $\mathrm{overlap}_{\min}$ and performance-based similarity $\rho_\text{Pearson}$ across 28 benchmark pairs is 68.4\% (Pearson) / 60.7\% (Spearman) (Table~\ref{tab:overlap_corr}).
While positive, this underestimates the true relationship.
A closer inspection reveals the source of the gap: benchmarks like GSM8K exhibit high performance-based similarity with many other benchmarks, even those with low feature overlap.
The reason is a confounding factor, namely general model ability: models trained longer tend to improve on all benchmarks simultaneously, inflating performance correlations even between unrelated benchmarks.
This ``rising tide'' effect creates spurious similarity that has nothing to do with shared capabilities.

\paragraph{Controlling for general ability.}
To isolate the capability-specific signal, we partial out MMLU, which serves as a proxy for general ability:
\begin{equation}
    \rho_{\text{partial}}(\mathcal{D}_i, \mathcal{D}_j \mid \mathcal{D}_{\text{MMLU}}) = \frac{\rho(\mathcal{D}_i, \mathcal{D}_j) - \rho(\mathcal{D}_i, \mathcal{D}_{\text{MMLU}}) \cdot \rho(\mathcal{D}_j, \mathcal{D}_{\text{MMLU}})}{\sqrt{1 - \rho(\mathcal{D}_i, \mathcal{D}_{\text{MMLU}})^2} \cdot \sqrt{1 - \rho(\mathcal{D}_j, \mathcal{D}_{\text{MMLU}})^2}}.
\end{equation}
After this correction, the partial Pearson correlation improves to \textbf{75.5\%} (Table~\ref{tab:overlap_corr}), providing evidence that feature overlap captures benchmark-specific capability similarity beyond general model quality.

\begin{table}[t]
    \centering
    \caption{Correlation between symmetric feature overlap ($\text{overlap}_\text{min}$) and performance-based similarity ($\rho_\text{Pearson}$) across 28 benchmark pairs, before and after controlling for general ability.}
    \label{tab:overlap_corr}
    \begin{tabular}{lcc}
        \toprule
        \textbf{Correlation metric} & \textbf{Direct} & \textbf{Partial (control: MMLU)} \\
        \midrule
        Pearson  & 68.4 & \textbf{75.5} \\
        Spearman & 60.7 & \textbf{71.3} \\
        \bottomrule
    \end{tabular}
\end{table}

\paragraph{Implications for evaluation suite design.}
This result has a direct practical implication: feature overlap can guide evaluation suite design without any model evaluation.
Benchmarks with low mutual overlap probe distinct capabilities and should both be retained; benchmarks with high overlap are candidates for consolidation.
For example, the asymmetric overlap analysis shows that $63\%$ of GSM8K's features are already covered by MATH, suggesting that an evaluation suite containing MATH can safely drop GSM8K with little loss of discriminative information.
Conversely, a benchmark (or data source) whose feature footprint has low overlap against all current suite members likely probes capabilities that are not yet covered.
In the language of out-of-distribution detection, such a benchmark is ``OOD'' with respect to the existing suite, making it a natural candidate for inclusion to close capability gaps.
\clearpage


\section{Application: Data Classification}
\label{sec:applications_classification}

\begin{figure}[h]
  \centering
  \includegraphics[width=1.0\linewidth]{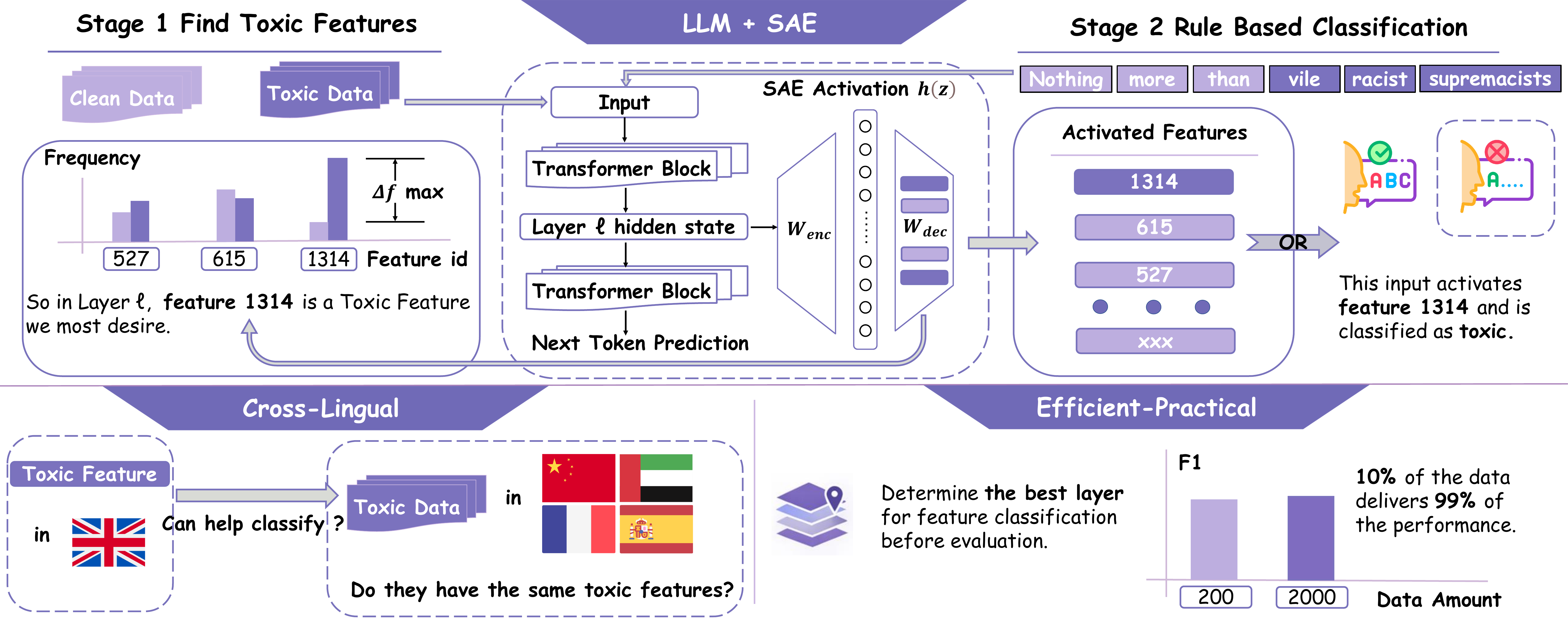}
    \vspace{-7mm}
    \caption{Overview of the SAE-based toxicity classification pipeline. Feature discovery is performed on a fixed selection split by measuring how often each SAE feature fires on toxic versus clean examples. The resulting features are then used directly as a rule-based classifier on held-out data.}
   \label{fig:sae_classifier_pipeline}
\end{figure}

A natural test of whether SAE features are useful in practice is to ask whether they can directly support a downstream classifier. We study this question on the multilingual toxicity corpus~\citep{textdetox}, and focus on a deliberately constrained setting: rather than training a new classification head, we ask whether a small set of SAE features can be used as the classifier itself. This framing is important. If the resulting classifier is effective, then SAE features are not merely descriptive tools for post hoc analysis; they are actionable variables that can support concrete prediction while preserving transparency.

Our results suggest that a small set of toxicity-biased SAE features already yields a strong rule-based classifier, despite using no additional supervised head and no gradient-based fitting after the SAE is fixed. The same features also reveal broader structure: some toxicity-related directions are shared across languages, some transfer surprisingly well from English to other languages, and the entire pipeline can be made substantially more efficient through simple layer selection and reduced feature-discovery data. Taken together, these results position SAE features as a practical interface between mechanistic interpretability and usable classification systems.

\subsection{SAE-Based Toxicity Classifier}
\label{subsec:sae_based_classifier}
We aim to keep the SAE-based classification method as simple as possible, since simplicity makes it easier to extend to practical applications. For each language, we identify SAE features that fire substantially more often on toxic examples than on clean ones, and use these features directly as detectors on held-out data. The resulting predictor is sparse, discrete, and easily interpretable: each positive prediction can be traced to a small set of latent features and the layer where they emerge.

The design avoids complex formulas to identify classification features and does not require interpreting them in advance. Once an SAE is available, it can be used directly for classification. This simplicity is key: the goal is not only to detect toxicity, but to keep the path from model internals to prediction transparent.

Figure~\ref{fig:sae_classifier_pipeline} highlights that the entire method reduces to a simple and transparent two-stage pipeline: discover toxic SAE features on a fixed selection split, then apply them directly as a rule-based classifier on held-out data. This decomposition is important for interpretation, because every prediction can be traced back to a feature, layer, and token position rather than to an opaque classification head.

\subsubsection{Toxic Feature Discovery}
\label{subsubsec:toxic_feature_discovery}

We study SAE-based toxicity classification on the multilingual toxicity dataset~\citep{textdetox}. Our experiments use Qwen3-1.7B and Qwen3-8B~\citep{qwen3} with their corresponding SAEs (32k and 64k). From the dataset, we retain \(13\) languages with \(5\)k examples each: English (en), Russian (ru), Ukrainian (uk), German (de), Spanish (es), Amharic (am), Chinese (zh), Arabic (ar), Hindi (hi), Italian (it), French (fr), Tatar (tt), and Japanese (ja). For each language, we start from a balanced pool of roughly 5{,}000 examples and keep a fixed, reproducible split: 4{,}000 examples for feature discovery (2{,}000 toxic and 2{,}000 clean) and 1{,}000 examples for evaluation (500 toxic and 500 clean).

Feature discovery is performed independently at each transformer layer. We run the input text through the base model in prefill mode, extract the residual stream at the target layer, and pass those activations through the corresponding SAE encoder.

Let \(a_{i,t,f}^{(\ell)}\) denote the activation of SAE feature \(f\) at token position \(t\) for example \(i\) at layer \(\ell\). We then convert token-level activations into an example-level binary firing variable:
\begin{equation}
h_{i,f}^{(\ell)} = \mathbb{1} \left[\max_t a_{i,t,f}^{(\ell)} > \epsilon \right],
\end{equation}
where \(\epsilon\) is a small threshold (set to \(0\) in our implementation). Intuitively, a feature is counted as firing on an example if it activates anywhere in the prompt.

Using these binary firing indicators, we compute how often each feature appears on toxic vs. clean data:
\begin{equation}
\Delta_f^{(\ell)} =
\Pr \left(h_{i,f}^{(\ell)}=1 \mid y_i=1\right)
-
\Pr \left(h_{i,f}^{(\ell)}=1 \mid y_i=0\right),
\end{equation}
where \(y_i=1\) denotes a toxic label and \(y_i=0\) a clean label. We then rank features by \(\Delta_f^{(\ell)}\) and select the top \(K\) features at each layer. This scoring rule is intentionally minimal: it favors features that are not merely active, but selectively active on toxic data.

This procedure gives the classifier an interpretable basis from the start. Each selected feature comes with a clear quantitative signature---its toxic firing frequency, its clean firing frequency, and their difference. The classifier is therefore built from features that are explicitly biased toward toxic data, rather than from an opaque learned boundary in a high-dimensional latent space.

\subsubsection{Rule-Based Classification with Selected Features}
\label{subsubsec:rule_based_classification}

Once a set of toxic-biased features has been selected, evaluation on the test split is straightforward. For a target layer \(\ell\), we again extract the residual stream, encode it with the SAE, and retain only the selected feature set \(S_\ell\). A test example is classified as toxic if any selected feature fires at any token position:
\begin{equation}
\hat{y}_i =
\mathbb{1} \left[
\max_{f \in S_\ell}\max_t a_{i,t,f}^{(\ell)} > \epsilon
\right].
\end{equation}
This is an OR-rule over a small number of latent features. No additional classifier head is trained, and no feature weights are learned after selection.

\begin{figure}[t]
  \centering
  \includegraphics[width=1.0\linewidth]{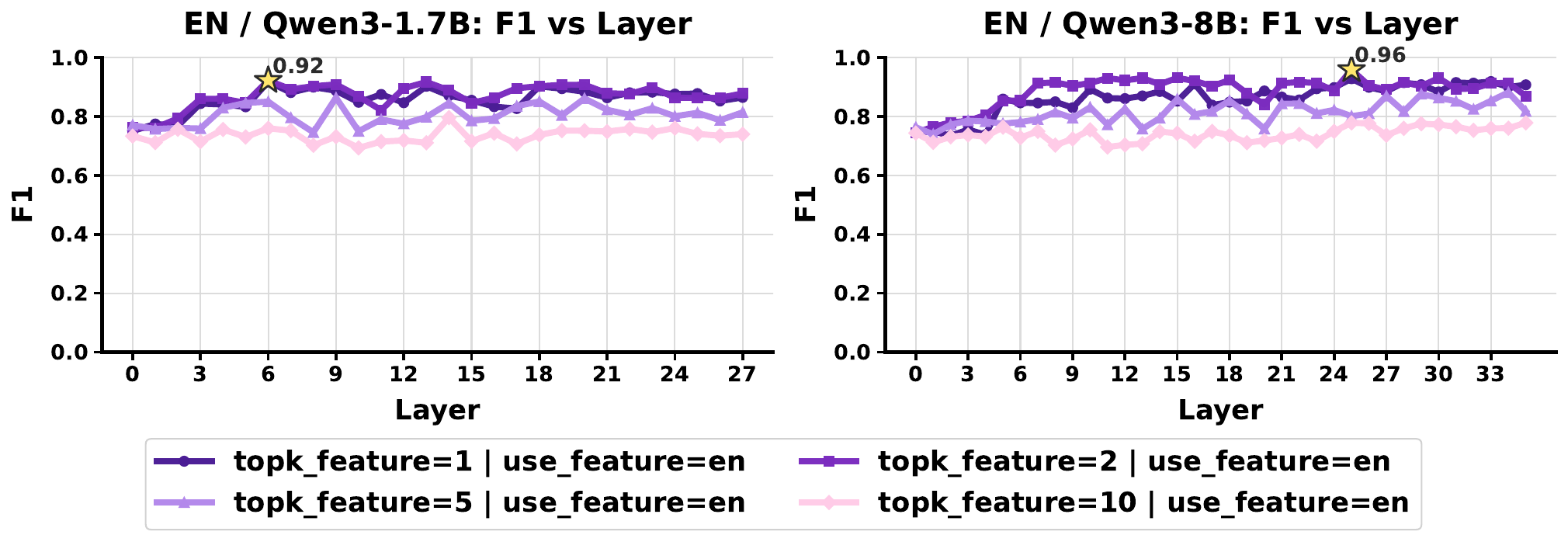}
  \vspace{-7mm}
  \caption{\textbf{Layer-wise F1 of the SAE-based toxicity classifier on English.} 
  \textbf{Left:} Qwen3-1.7B.\textbf{ Right:} Qwen3-8B. The curves report held-out F1 across layers using top-$K$ toxic-biased SAE features discovered in English, with $K \in \{1,2,5,10\}$. Star markers indicate the best F1 layer for each model. \textbf{Without training any additional classifier, the existing SAE can be used directly for classification, achieving an F1 score above 0.90 for identifying toxic features in English text.}}
  \label{fig:sae_classifier_f1_en}
  \vspace{-4mm}
\end{figure}

From Figure~\ref{fig:sae_classifier_f1_en}, we can see that a set of SAE features already yields a highly effective English toxicity classifier, with best held-out F1 exceeding 0.90 in both models. The strongest performance is concentrated in a relatively narrow band of middle-to-late layers, and increasing $K$ beyond a very small value brings limited additional benefit. This indicates that the toxicity signal is sparse and concentrated in a handful of highly selective latent features.

Strong classification performance is achieved with only a small number of identifiable features, rather than a dense combination of many latent dimensions. The decision rule also remains local and interpretable: each positive prediction can be traced to the feature, layer, and token position that triggered it, a level of transparency that is difficult to obtain with a trained classification head.

\subsection{Cross-Lingual Generalization of Toxic Features}
\label{subsec:cross_lingual_generalization}

A strong single-language classifier is useful, but it leaves a deeper question: are the discovered features capturing language-specific lexical cues, or more abstract structures associated with toxic intent? The multilingual setting provides a way to test this. We therefore examine both the overlap of discovered features across languages and the transfer performance of features discovered in English.

The answer is mixed, but encouraging. Toxicity-related SAE features are neither fully language-agnostic nor purely language-specific. Instead, the results suggest a layered structure. Some features are shared across languages, particularly in the middle layers, and this shared structure is sufficient to support meaningful cross-lingual transfer.

\subsubsection{Shared Toxic Structure Across Languages}
\label{subsubsec:shared_toxic_structure}
\begin{figure}[t]
  \centering
  \includegraphics[width=1.0\linewidth]{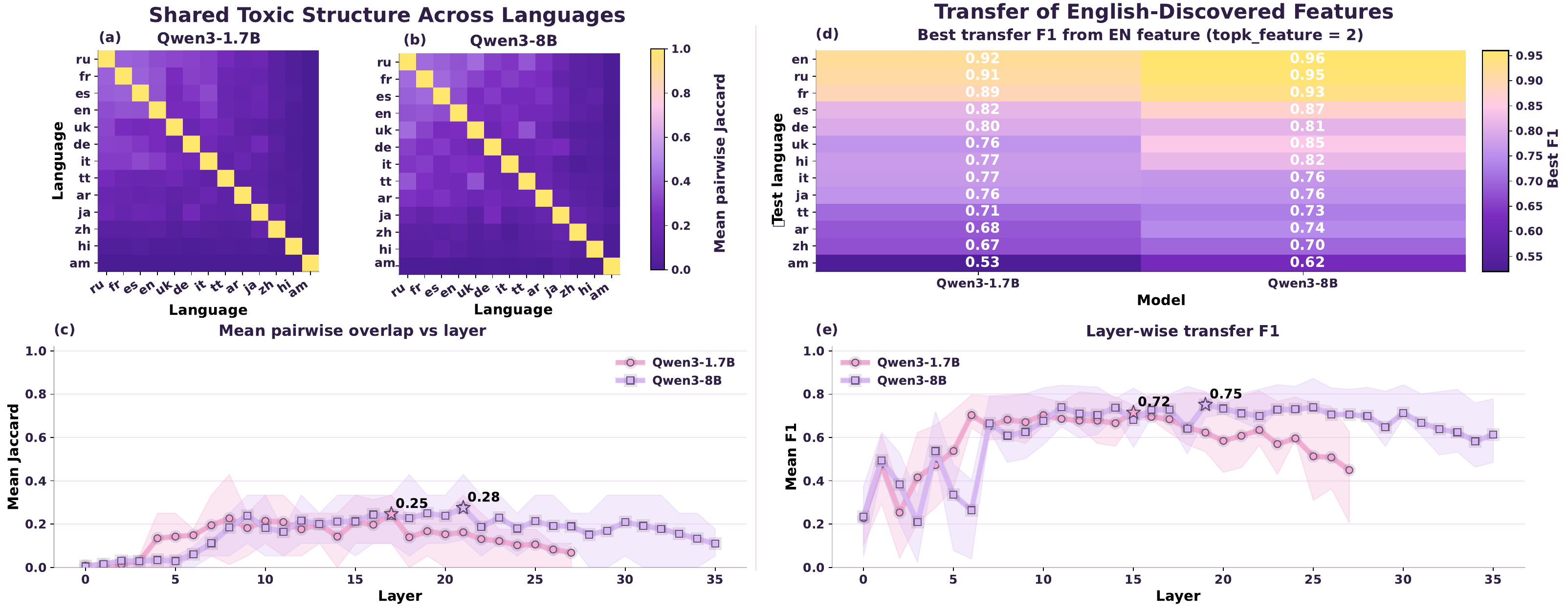}
  \vspace{-7mm}
  \caption{\textbf{Cross-lingual structure and transfer of toxic SAE features.}
   \textbf{Panel (a)} shows the overlap of top-10 toxic SAE features in Qwen3-1.7B, and \textbf{panel (b)} shows the same for Qwen3-8B. \textbf{Panel (c)} shows the layer-wise mean overlap, with shaded bands indicating the interquartile range (IQR) across language pairs. \textbf{Panel (d)} shows the best held-out F1 for each test language using features discovered in English. \textbf{Panel (e)} shows the layer-wise mean transfer F1, with shaded bands indicating the interquartile range (IQR) across languages. Star markers indicate the best layer for each model. \textbf{Toxic SAE features show structured cross-lingual sharing, and English-discovered features transfer well to many languages, especially in larger models.}}
  \label{fig:cross-lingual}
  \vspace{-4mm}
\end{figure}
We first ask whether the toxic SAE features discovered in different languages are, in fact, capturing related internal structure. To test this, we measure the overlap between the top toxic feature sets discovered independently in each language. At a fixed layer, we compute the Jaccard overlap between the top-$K$ feature indices for every language pair, and then examine how this overlap varies across both language pairs and layers.

Several trends emerge from Figure~\ref{fig:cross-lingual}. Cross-lingual sharing is clearly present, but it is uneven across language pairs. Panels (a) to (c) show that overlap is highest for typologically closer languages, especially among European languages, and substantially weaker for more distant pairs. This suggests that toxicity is not represented in a fully language-agnostic feature basis, and that linguistic distance remains an important factor in which features are recovered.

The layer pattern is equally informative. Shared structure is most pronounced in the middle layers, rather than at the bottom or top of the network, which suggests that these layers provide the clearest substrate for multilingual toxic feature discovery. The same pattern appears in both Qwen3-1.7B and Qwen3-8B, with the larger model showing somewhat stronger and more stable overlap overall. Taken together, these results suggest that toxicity-related SAE features are not identical across languages, but are consistent enough to motivate direct transfer experiments.

\subsubsection{Transfer of English-Discovered Features}
\label{subsubsec:transfer_english_features}

Overlap alone does not show whether a feature set discovered in one language can be used directly for classification in another. We therefore consider a stricter test: discover toxic features in English, then apply those same features to held-out data in other languages without rediscovering them. This directly tests whether English-discovered SAE features capture portable toxicity-related structure rather than language-specific lexical cues.

The transfer results are encouraging, but clearly uneven. Panels (d) and (e) of Figure~\ref{fig:cross-lingual} show strong transfer to English itself and to several European languages, including Russian and French, while more distant languages such as Arabic, Chinese, and especially Amharic remain substantially harder. Cross-lingual transfer is therefore graded rather than uniform: performance declines with linguistic distance, but remains useful across a broad set of languages.


Scaling to Qwen3-8B improves both the level and stability of cross-lingual transfer, with optimal layers shifting deeper. This suggests that SAE-based toxicity detectors discovered in English can serve as effective starting points for multilingual detection without full rediscovery, particularly in larger models.

\begin{figure}[t]
  \centering
  \includegraphics[width=1.0\linewidth]{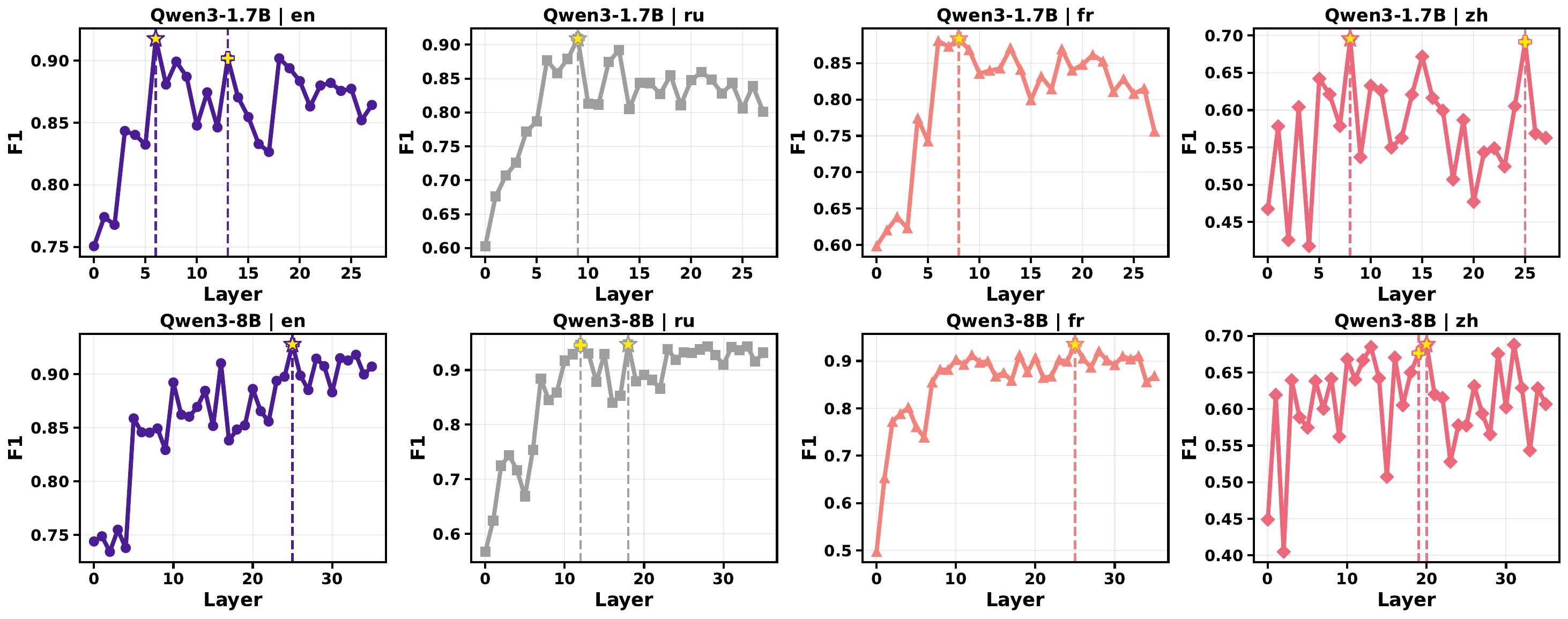}
  \vspace{-7mm}
  \caption{\textbf{A proxy for selecting high-performing layers before evaluation.}
  Row 1 shows Qwen3-1.7B, and Row 2 shows Qwen3-8B; columns correspond to English, Russian, French, and Chinese. In each subplot, the curve shows held-out F1 over layers, the yellow star marks the best evaluation layer, and the yellow cross marks the layer whose strongest discovered feature most clearly separates toxic from clean examples. \textbf{top1-diff is a reliable proxy for evaluation free layer selection while retaining nearly all achievable performance.}}
  \label{fig:step2_f1_vs_layer_top1_delta_proxy}
  \vspace{-4mm}
\end{figure}

\subsection{Toward Efficient and Practical Classification}
\label{subsec:efficient_practical_classification}

The results above establish that SAE features can already support accurate, interpretable toxicity classification with meaningful cross-lingual transfer. We next ask whether the same approach can be made both simpler and stronger in practice:

\begin{quote}
\itshape
Can we identify the right layer before evaluation, and can a small combination of layers improve on the best single-layer detector? (Section~\ref{subsubsec:layer_selection_multilayer})
\end{quote}

\begin{quote}
\itshape
Can the data and computation required for SAE-based feature discovery help explain why we use it rather than train an additional classifier (Section~\ref{subsubsec:data_efficiency})
\end{quote}

\subsubsection{Layer Selection and Multi-Layer Composition}
\label{subsubsec:layer_selection_multilayer}

Our starting point is simple: if a layer contains even one feature that separates toxic from clean examples especially well during feature discovery, that layer is likely to be useful at test time. We therefore use the strongest toxic clean frequency gap in a layer as a simple proxy for layer quality, which we call \emph{top1-diff}:
\begin{equation}
d^{(\ell)} = \max_f \Delta_f^{(\ell)},
\qquad
\ell^\star = \arg\max_{\ell} d^{(\ell)},
\end{equation}
where $d^{(\ell)}$ is the \emph{top1-diff} score of layer $\ell$. We then select the layer with the largest \emph{top1-diff} and use the feature set discovered at that layer for classification, exactly as in Section~\ref{subsec:sae_based_classifier}. This provides an evaluation-free proxy for layer quality before running a full sweep over held-out performance.

Figure~\ref{fig:step2_f1_vs_layer_top1_delta_proxy} makes the main point clear: the layer selected by \emph{top1-diff} is usually the best layer or very close to it. This holds across languages and across both model sizes, which means that much of the cost of a full layer sweep can be avoided with a simple statistic computed during feature discovery.

We can then extend the same idea to a multi-layer composition classifier. We rank layers by their \emph{top1-diff} scores, retain the top $m$ layers, and keep only the single best feature from each selected layer:
\begin{equation}
f_\ell^\star = \arg\max_f \Delta_f^{(\ell)},
\qquad
\hat{y}_i =
\mathbb{1} \left[
\max_{\ell \in \mathcal{L}_{\mathrm{top}}}\max_t a_{i,t,f_\ell^\star}^{(\ell)} > \epsilon
\right].
\end{equation}
The motivation is equally simple: when no single layer contains a dominant toxicity signal, several moderately useful layers may together provide a stronger detector. This keeps the classifier sparse and inspectable, since each positive prediction can still be traced to a small set of explicit features.

\begin{figure}[t]
    \centering
    \includegraphics[width=1.0\linewidth]{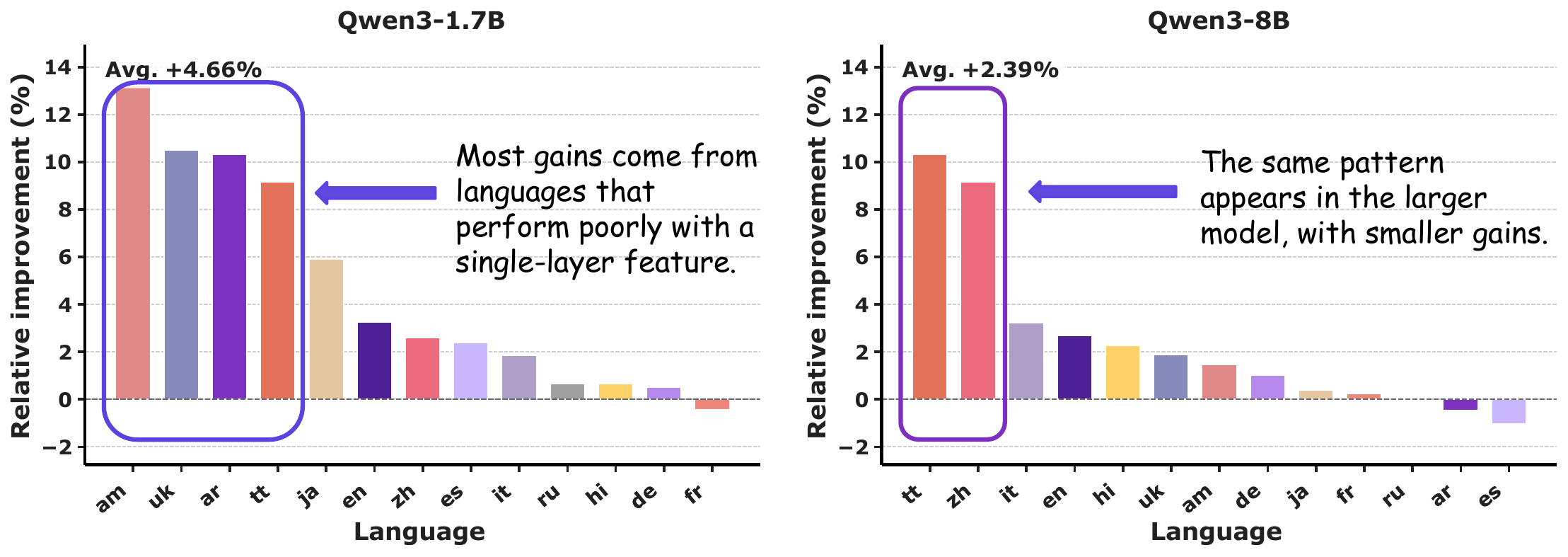}
    \vspace{-7mm}
    \caption{\textbf{Relative improvement from multi-layer composition.}
    Left: Qwen3-1.7B. Right: Qwen3-8B. For each language, the bar shows the relative improvement of the best multi-layer classifier over the single-layer baseline, where layers are ranked by the top1-diff and a small number of top-ranked layers are combined. \textbf{Multi-layer composition is most useful as a targeted robustness mechanism, improving harder languages while preserving a sparse and interpretable classifier.}}
    \label{fig:multilayer_composition_improvement}
    \vspace{-2mm}
\end{figure}

The central message of Figure~\ref{fig:multilayer_composition_improvement} is that multi-layer composition is most useful when single-layer evidence is weak. We see that harder cases often improve more noticeably. The resulting recipe is straightforward: first rank layers by \emph{top1-diff}, use the best layer when its signal is already strong, and add a small number of top-ranked layers only when extra robustness is needed.

\subsubsection{Data Efficiency of Feature Discovery}
\label{subsubsec:data_efficiency}

An effective SAE-based classifier should not require a large feature discovery dataset to function. Taking full advantage of the general SAE, we want to know: how much downstream classification performance can be preserved when using a smaller discovery dataset?

\begin{figure}[t]
  \centering
  \includegraphics[width=1.0\linewidth]{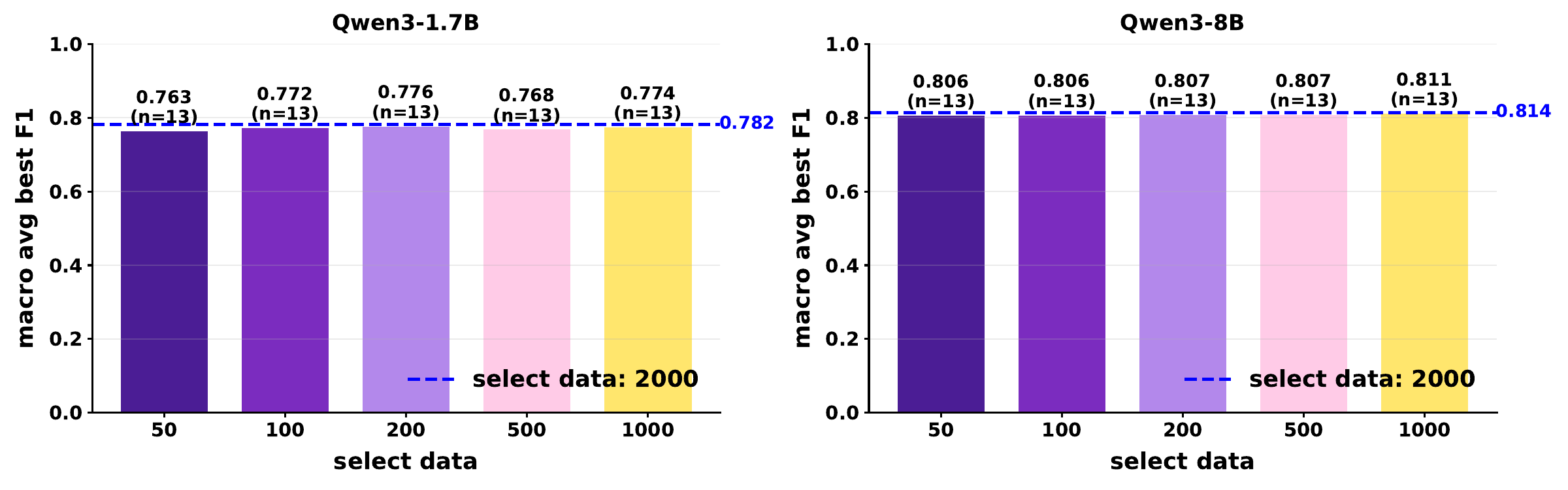}
  \vspace{-8mm}
  \caption{\textbf{Macro-average best F1 across languages under different toxic-feature selection sizes.}
  \textbf{Left:} Qwen3-1.7B. \textbf{Right:} Qwen3-8B. For each toxic-feature selection size, each bar reports the macro-average of the \emph{best} held-out F1 over layers, computed across 13 languages. The dashed blue line denotes the baseline setting with \textit{select data = 2000}. \textbf{For both models, using only 10\% of the original toxic-feature data achieves 99\% of the original classification performance.}}
  \label{fig:data_advantage}
  \vspace{-2mm}
\end{figure}

Figure~\ref{fig:data_advantage} shows that the classifier remains strong even with far less labeled data. This explains why we use SAE features directly for classification: once a good SAE is available, a small labeled set suffices to identify toxic features and build an effective, interpretable detector.

In particular, using only \(10\%\) of the original discovery data already recovers about \(99\%\) of the original classification performance, which means feature discovery is highly data-efficient. As the discovery budget grows, overlap with the full-data feature set rises quickly, which suggests that the most stable toxic-biased features are found early. More importantly, downstream performance remains close to the full-data baseline even when the discovery set is much smaller. 
\clearpage

\section{Application: Data Synthesis}
\label{sec:applications_synthesis}

\begin{figure}[h]
  \centering
  \includegraphics[width=1.0\linewidth]{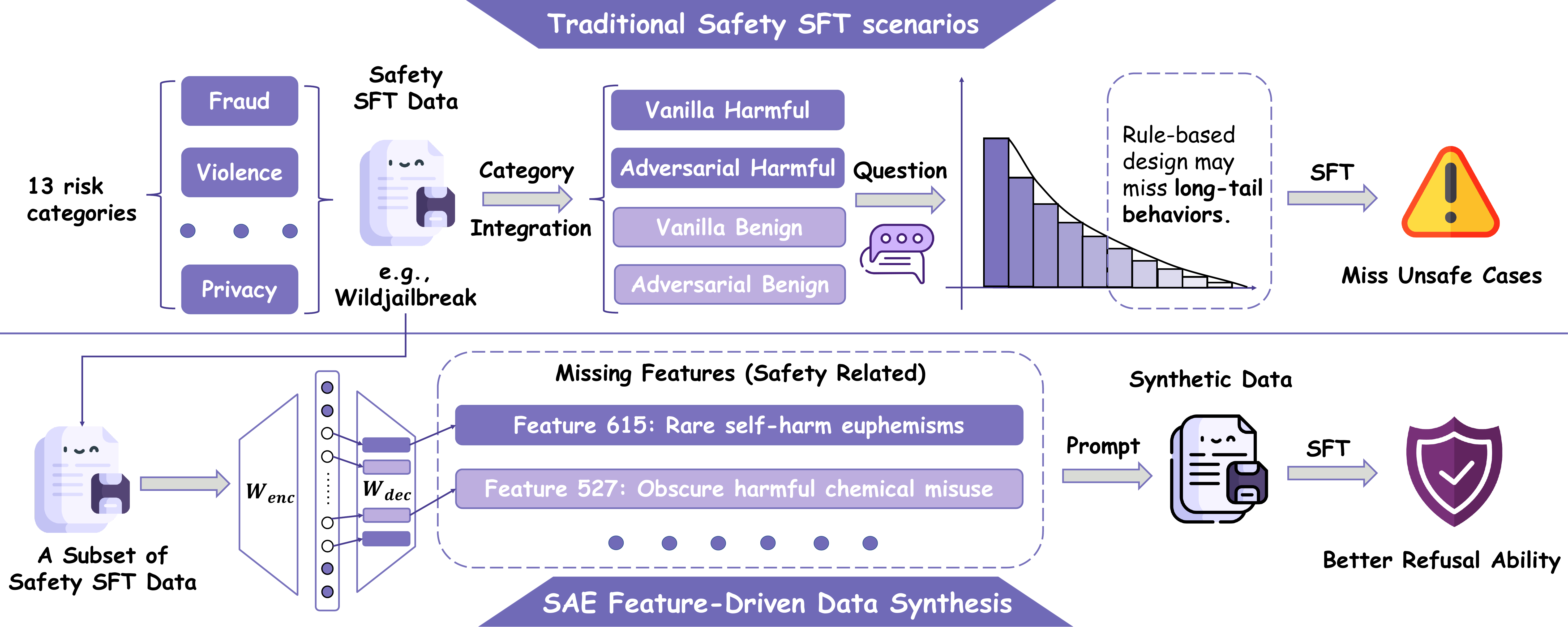}
    \caption{\textbf{Overview of the SAE-feature-driven safety data synthesis pipeline.}
    \textbf{Top:} Conventional safety SFT data, built from broad human-designed safety categories, can miss long-tail unsafe behaviors, so post-training improves refusal mainly for covered cases. \textbf{Bottom:} We pass safety SFT data through SAE to identify safety-relevant features that are missing, then use these features as synthesis targets. The resulting synthetic data are added back to the original safety SFT pool, enabling post-training to improve refusal coverage in long-tail scenarios.}
    \label{fig:feature_driven_synthesis_pipeline}
\end{figure}

Having demonstrated the value of SAE in data classification, we next turn to another data-centric direction: data synthesis. Recent work argues that refusal is not learned as a wholly new capability during post-training; instead, post-training links an already represented concept of harmful content to a specific action policy~\citep{lindsey2025biology}. In practice, however, safety SFT data are hard to scale to the full range of safety-relevant situations. Many important behaviors lie in the long tail, where natural sampling is either inefficient or prone to bias and noise.

Let's rethink what SAE features really represent: trained on data from the same distributional regime as the base model, SAEs encode many concepts learned during pretraining. Recent work shows that their value lies not in expanding SFT to cover the full pretraining distribution, but in exposing concepts the model knows without turning them into reliable behavior~\citep{li2026less}. With limited data, feature-driven synthesis can target these gaps and teach missing safety behaviors more efficiently.

Under this view, the role of SAE-guided synthesis is not to recreate the full pretraining distribution, but to identify and reinforce concepts the model already knows but has not yet turned into reliable post-training behavior.

\subsection{Feature-Driven Safety Data Synthesis}
\label{subsec:feature_driven_safety_data_synthesis}

The central idea is to move data construction from the corpus level to the representation level. Instead of asking only which safety prompts to sample, we first identify safety-relevant SAE features that are missing or weakly covered, and then synthesize prompt-completion pairs that are explicitly designed to activate them. The resulting pipeline is simple: select target features, generate examples from their descriptions, and retain only those examples that are verified to hit the intended internal directions.

\subsubsection{Target Feature Discovery}
\label{subsubsec:synthesis_target_feature_discovery}

The first question is which internal safety directions should be strengthened before synthesizing any new data. Directly enumerating the full long tail of safety-relevant situations is difficult, so we begin from a smaller seed corpus drawn from the available safety supervision pool, denoted by \(D_{\mathrm{seed}}\). This seed corpus is used as a diagnostic probe rather than as an exhaustive description of the safety space. Its purpose is to tell us which safety-relevant SAE features are already reached by existing supervision and which ones remain absent or only weakly supported.

As in Section~\ref{subsec:sae_based_classifier}, let \(a_{i,t,f}^{(\ell)}\) denote the token level activation of feature \(f\) at token position \(t\) for example \(i\) at layer \(\ell\), and let \(h_{i,f}^{(\ell)}\) denote the corresponding example level firing indicator. We first define a binary feature coverage variable:
\begin{equation}
c_f^{(\ell)}(D_{\mathrm{seed}}) = \mathbb{1} \left[\exists i \in D_{\mathrm{seed}}~~~\text{s.t.}~~~h_{i,f}^{(\ell)} = 1\right].
\end{equation}

This quantity indicates whether feature \(f\) is activated by at least one example in the seed corpus at layer \(\ell\). In other words, coverage is defined in feature space rather than prompt space. If
\(c_f^{(\ell)}(D_{\mathrm{seed}})=0\),
then the current supervision never reaches that internal direction. If
\(c_f^{(\ell)}(D_{\mathrm{seed}})=1\),
then the feature is at least touched somewhere in the seed corpus. This notion is intentionally coarse. It does not measure how often a feature appears or how strongly it is activated. It only asks whether the current supervision reaches that feature at all. For this reason, coverage should be understood as a first pass support estimate over the feature inventory rather than a complete measure of training adequacy.

Coverage alone is not sufficient for target selection, because not every uncovered feature is necessarily useful for safety post-training. We therefore combine this support signal with a semantic-relevance filter. Each feature is paired with a natural-language explanation, and a judge model assigns a relevance score \(s_f^{(\ell)} \in [0,1]\) that estimates whether the feature corresponds to behavior that is useful for safety supervised fine-tuning. These explanations can be obtained from top-activating contexts or from an automatic feature-interpretation pipeline~\citep{autointerp}. The judge is used only to filter and rank candidate features for synthesis; it does not directly determine whether a generated example is retained. Retention is decided by the representation-level verification step described below. We then define the candidate target inventory as
\begin{equation}
\mathcal{T} = \left\{(\ell,f)~:~s_f^{(\ell)} \ge \tau\right\},
\end{equation}
where \(\tau\) is a confidence threshold.

In practice, the highest-priority synthesis targets are the eligible features that are not covered by the seed corpus:
\begin{equation}
\mathcal{T}_{\mathrm{miss}}(D_{\mathrm{seed}})
=
\left\{
(\ell,f) \in \mathcal{T}
~:~
c_f^{(\ell)}(D_{\mathrm{seed}})=0
\right\}.
\end{equation}
When a larger synthesis budget is available, this set can be further expanded to include weakly covered features, for example features whose firing frequency on \(D_{\mathrm{seed}}\) is nonzero but below a small support threshold. This distinction separates \emph{semantic eligibility}, determined by \(s_f^{(\ell)}\), from \emph{coverage priority}, determined by the seed corpus.

Under this formulation, semantic relevance determines which features are eligible targets, while coverage determines how those targets are prioritized. Features in \(\mathcal{T}_{\mathrm{miss}}(D_{\mathrm{seed}})\) are natural synthesis targets because they are safety-relevant but completely absent from the current supervision. Features in \(\mathcal{T}\setminus \mathcal{T}_{\mathrm{miss}}(D_{\mathrm{seed}})\) may also remain useful targets if they are safety-critical yet appear only sparsely or weakly in the seed corpus. In this sense, target discovery is driven by feature semantics and informed by feature coverage: instead of asking which prompts are missing from the dataset, we ask which internal safety-relevant directions are not yet adequately supported by the current data.

\subsubsection{Data Synthesis from Feature Descriptions}
\label{subsubsec:feature_description_guided_synthesis}

Once a target feature \((\ell,f) \in \mathcal{T}\) has been selected, the next step is to convert that feature level target into concrete supervision. Each target feature is paired with a natural language explanation \(e_f^{(\ell)}\), and we use this explanation as the starting point for data construction. The goal is not to reproduce prompts already present in the corpus, but to generate examples that express the behavior encoded by the target feature and can therefore strengthen that internal direction during post-training.

Our synthesis pipeline has three stages: prompt construction, response construction, and representation level verification. Prompt construction determines what kind of request should be expressed. Response construction determines the desired model behavior for that request. Verification checks whether the resulting example actually activates the intended feature. This separation makes the pipeline both interpretable and controllable.

\paragraph{Prompt construction.}
For each target feature, we first generate a \emph{vanilla} prompt
\(
x_{\ell,f}^{\mathrm{van}}
\)
that expresses the underlying intent in a direct and natural form. We then construct one or more adversarial variants
\(
\{x_{\ell,f,k}^{\mathrm{adv}}\}_k
\)
that preserve the same core intent while changing the surface form to resemble more realistic jailbreak-style inputs. Formally, we write
\begin{equation}
x_{\ell,f}^{\mathrm{van}}
\sim
G_{\mathrm{van}} \left(e_f^{(\ell)}\right),
\qquad
x_{\ell,f,k}^{\mathrm{adv}}
\sim
G_{\mathrm{adv}} \left(x_{\ell,f}^{\mathrm{van}}, \eta_k\right),
\end{equation}
where \(G_{\mathrm{van}}\) maps a feature explanation to a canonical request, \(G_{\mathrm{adv}}\) rewrites that request into a more adversarial form, and \(\eta_k\) indexes different attack styles. The vanilla prompt serves as a clean semantic anchor, while the adversarial variants broaden coverage toward forms that are more likely to appear in practice.

\paragraph{Response construction.}
The safety label \(z\) is assigned according to the risk category expressed by the target feature and the generated prompt. This label determines whether the desired completion should refuse the request or answer it normally. Given a prompt \(x\) and a safety label
\(
z \in \{\text{harmful}, \text{benign}\},
\)
we generate a response
\begin{equation}
y \sim G_{\mathrm{resp}}(x, z).
\end{equation}
When \(z=\text{harmful}\), the target response is a refusal-style completion that declines the request and, when appropriate, redirects to a safe alternative. When \(z=\text{benign}\), the target response is a normal helpful completion. This distinction is essential: the aim of safety fine-tuning is not to suppress broad regions of behavior, but to sharpen the boundary between harmful and benign requests.

\paragraph{Representation level verification.}
Prompt intent alone is not enough to guarantee that a synthesized example actually targets the desired internal direction. We therefore verify each synthesized prompt in feature space. For a candidate example \(i\), we retain it for target \((\ell,f)\) only if its example level firing indicator satisfies \(h_{i,f}^{(\ell)} = 1\), meaning that the example activates the target feature at the source layer. In practice, adversarial rewrites may also be filtered before this step to preserve semantic equivalence and risk category. The key point is that examples are not accepted solely because they look relevant at the text level; they must also be validated at the representation level.

This verification step gives the method its main advantage. The synthesis target is specified in feature space, and the final data are also selected in feature space. As a result, the generated corpus is aligned not only with textual descriptions of safety-relevant behavior, but also with the internal directions that the model is expected to strengthen during post-training.

To summarize how well a synthetic dataset \(D\) covers the target inventory, we define the target feature coverage as
\begin{equation}
\mathrm{Cov}(D)
=
\frac{1}{|\mathcal{T}|}
\sum_{(\ell,f)\in\mathcal{T}}
\mathbb{1} \left[
\exists i \in D~~~\text{s.t.}~~~
h_{i,f}^{(\ell)} = 1
\right].
\end{equation}
This quantity measures the fraction of target features that are activated by at least one retained example in the synthetic dataset. A target feature is counted as covered if the dataset contains at least one example that reaches that feature at the corresponding layer. Coverage is therefore defined at the level of internal representations rather than prompt categories. A synthetic dataset achieves high coverage when it reaches a large portion of the target feature set, not merely when it contains many superficially diverse prompts.

Under this formulation, feature-driven synthesis is more than prompt generation from textual descriptions. It is a representation-aware data construction procedure: feature explanations define what to generate, and feature activations determine what to keep.

\subsection{Toward Controllable Safety Post-Training}
\label{subsec:toward_controllable_safety_post_training}

We next ask whether the above approach is useful in practice along two dimensions:

\begin{quote}
\itshape
Can feature-driven synthesis cover safety-relevant SAE features more efficiently than natural sampling or unconstrained safety-related synthesis? (Section~\ref{subsubsec:feature_coverage})
\end{quote}

\begin{quote}
\itshape
Does this improved feature coverage translate into a better safety--utility tradeoff after SFT? (Section~\ref{subsubsec:safety_utility_tradeoff})
\end{quote}

\begin{figure}[t]
  \centering
  \includegraphics[width=1.0\linewidth]{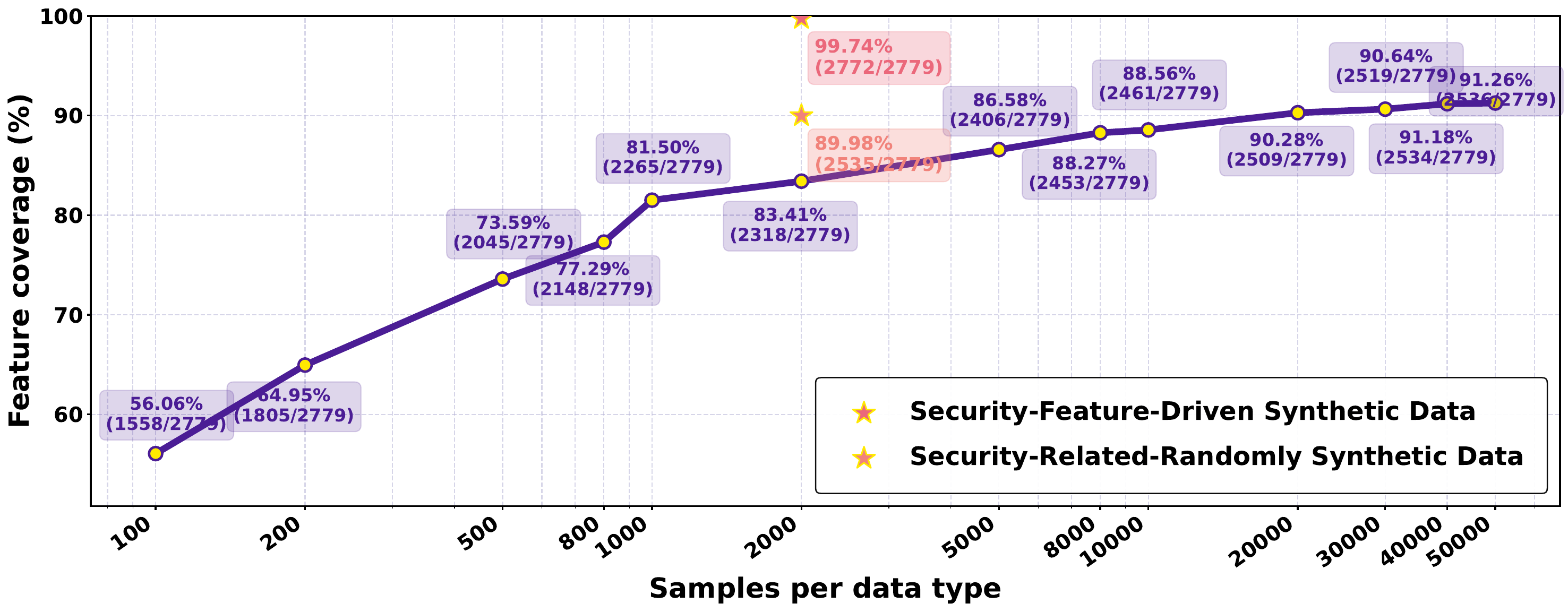}
    \caption{\textbf{Coverage of target safety features under different data construction strategies.}
    The curve shows how target-feature coverage grows as we increase the number of naturally sampled examples in each data category. Star markers indicate two matched-budget synthetic alternatives. \textbf{Feature-driven synthesis nearly saturates the target feature set with a small budget, while natural sampling and random security-related synthesis leave substantial gaps.}}
  \label{fig:feature_driven_synthesis_coverage}
\end{figure}

\subsubsection{Training and Evaluation Setup}
\label{subsubsec:synthesis_training_and_evaluation}

Our base model is Qwen3-8B~\citep{qwen3}. All synthesis targets are defined with respect to an SAE trained on its layer-\(30\) residual stream, with a latent dimensionality of approximately \(65\)k. For target discovery and synthesis, we draw on the WildJailbreak training corpus~\citep{wildjailbreak}, which contains four complementary data types: vanilla harmful, vanilla benign, adversarial harmful, and adversarial benign prompts. We follow the original data-construction recipe closely: prompts are generated with GPT-4~\citep{gpt4}, and responses are generated mainly with GPT-3.5.

For the coverage analysis in Section~\ref{subsubsec:feature_coverage}, target features are identified from a stratified seed set drawn from this WildJailbreak training corpus. We use its mixture of direct and adversarial, harmful and benign examples as the seed distribution for discovering safety-relevant features and for measuring how efficiently different data-construction strategies cover them.

For the downstream SFT results in Section~\ref{subsubsec:safety_utility_tradeoff}, we combine three data sources: general instruction data from Alpaca~\citep{alpaca}, real safety data from WildJailbreak, and synthetic safety data produced by our pipeline. We fine-tune the model with LoRA~\citep{lora}, keeping the training mixture balanced across harmful and benign examples as well as across the different safety data categories. The key comparison keeps the total safety-data budget fixed and replaces random synthetic safety data with feature-driven synthetic data. Safety is evaluated on harmful and benign prompts from the WildJailbreak test set, and general capability is evaluated on IFEval~\citep{ifeval}, TruthfulQA~\citep{truthfulqa}, MMLU~\citep{mmlu}, GSM8K~\citep{gsm8k}, and BBH~\citep{bbh}.

\subsubsection{Coverage Efficiency of Feature-Driven Synthesis}
\label{subsubsec:feature_coverage}

We first ask whether feature-driven synthesis covers the target inventory more efficiently than alternative data construction strategies. Figure~\ref{fig:feature_driven_synthesis_coverage} compares three settings: natural sampling from the safety corpus, random safety-related synthesis, and feature-driven synthesis. Coverage is measured by \(\mathrm{Cov}(D)\) from Section~\ref{subsubsec:feature_description_guided_synthesis}.

The result is straightforward. Natural sampling improves coverage only gradually, especially once the remaining targets move deeper into the long tail. Random safety-related synthesis improves coverage to some extent, but still leaves a substantial portion of the target inventory uncovered. Feature-driven synthesis is different: under the same matched budget, it reaches 99.74\% coverage and nearly saturates the target set.

This is the central empirical advantage of the method. Natural sampling depends on whether rare safety patterns happen to appear, and unconstrained synthesis can still miss the internal directions that matter most for post-training. Feature-driven synthesis instead targets those directions explicitly and verifies afterward that they were actually activated.

\subsubsection{Results with Synthetic Data}
\label{subsubsec:safety_utility_tradeoff}

\begin{table*}[t]
\centering
\caption{\textbf{Safety and capability results with feature-driven synthetic data.}
ASR, RR, and Acc measure safety behavior; IFEval, TruthfulQA, and MMLU measure general capability; GSM8K and BBH measure reasoning. Best results are in bold and second-best results are underlined, with ties all highlighted. Within the safety metrics, we highlight only Acc. \textbf{Adding \(4\)k feature-driven synthetic examples to \(4\)k real safety examples already approaches the effect of much larger safety-only SFT mixtures.}}
\small
\setlength{\tabcolsep}{4pt}
\resizebox{\textwidth}{!}{
\begin{tabular}{>{\raggedright\arraybackslash}p{4.5cm} c c c c c c c c}
\toprule
& \multicolumn{3}{>{\columncolor{safetygroup}}c}{\textbf{Safety}}
& \multicolumn{3}{>{\columncolor{generalgroup}}c}{\textbf{General}}
& \multicolumn{2}{>{\columncolor{reasongroup}}c}{\textbf{Reasoning}} \\
\cmidrule(lr){2-4} \cmidrule(lr){5-7} \cmidrule(lr){8-9}
\textbf{SFT training data}
& \cellcolor{safetygroup}\textbf{ASR\(\downarrow\)}
& \cellcolor{safetygroup}\textbf{RR\(\downarrow\)}
& \cellcolor{safetygroup}\textbf{Acc\(\uparrow\)}
& \cellcolor{generalgroup}\textbf{IFEval\(\uparrow\)}
& \cellcolor{generalgroup}\textbf{TruthfulQA\(\uparrow\)}
& \cellcolor{generalgroup}\textbf{MMLU\(\uparrow\)}
& \cellcolor{reasongroup}\textbf{GSM8K\(\uparrow\)}
& \cellcolor{reasongroup}\textbf{BBH\(\uparrow\)} \\
\midrule

\multicolumn{9}{c}{\textit{Trained on general SFT data only}} \\
\midrule
Alpaca \(50\)k~\citep{alpaca}
& 73.0 & 3.5 & 61.75
& 51.94 & 56.80 & \textbf{76.58}
& 79.00 & 76.73 \\
\midrule

\multicolumn{9}{c}{\textit{Trained on general SFT data + safety SFT data}} \\
\midrule
+ Safety \(8\)k~\citep{wildjailbreak}
& 22.0 & 34.5 & 71.75
& \underline{53.05} & \underline{57.11} & 76.25
& 73.71 & 76.79 \\
+ Safety \(40\)k
& 16.5 & 43.0 & 70.25
& 47.50 & 55.57 & 76.08
& 76.12 & \textbf{76.95} \\
+ Safety \(120\)k
& 21.0 & 21.5 & \textbf{78.75}
& 48.06 & 54.80 & \underline{76.34}
& \underline{82.56} & 76.29 \\
+ Safety \(200\)k
& 24.0 & 19.0 & \underline{78.50}
& 47.50 & 56.00 & 76.00
& \textbf{82.71} & 76.71 \\
\midrule

\multicolumn{9}{c}{\textit{Trained on general SFT data + safety SFT data + SAE synthetic data}} \\
\midrule
\rowcolor{randomsynthrow}
+ Safety \(4\)k + Random synth \(4\)k
& 20.0 & 36.0 & 72.00
& 48.98 & 56.94 & 76.08
& 74.45 & \underline{76.90} \\
\rowcolor{featuresynthrow}
+ Safety \(4\)k + Feature synth \(4\)k
& 24.0 & 20.5 & 77.75
& \textbf{53.23} & \textbf{57.32} & \textbf{76.58}
& 77.03 & 76.53 \\
\bottomrule
\end{tabular}
}

\label{tab:feature_driven_synthesis_results}
\end{table*}

The coverage results above establish that feature-driven synthesis is effective at the representation level. The remaining question is whether the coverage gains in Figure~\ref{fig:feature_driven_synthesis_coverage} carry through to downstream post-training behavior. Table~\ref{tab:feature_driven_synthesis_results} shows that targeting the right internal directions improves downstream safety while preserving, and in some cases improving, general utility. As a robustness check, we also use Gemini-3-Flash~\citep{gemini} for both prompt and response generation. The resulting performance is very close to that of the main setup, suggesting that the gain is driven by feature-targeted data construction itself rather than by the particular choice of generation models.

With only 8k total safety-related examples, feature-driven synthesis approaches the performance of the 120k safety-only setting. Concretely, using \(4\)k real safety examples together with \(4\)k feature-driven synthetic examples yields an overall safety accuracy of \(77.75\), compared with \(71.75\) for natural sampling at the same 8k budget. Notably, it also achieves the strongest IFEval and TruthfulQA scores in the table, indicating that targeted safety synthesis can improve safety without sacrificing general utility.

More importantly, the gain comes from targeted synthesis rather than synthetic data alone. This is clear in the matched comparison with unconstrained safety-related synthesis. Replacing \(4\)k random synthetic examples with \(4\)k feature-driven synthetic examples raises safety accuracy from \(72.00\) to \(77.75\), while also improving IFEval, TruthfulQA, MMLU, and GSM8K. Taken together, these results show that feature-driven synthesis improves the safety-utility tradeoff under a fixed data budget by making supervision more targeted rather than simply more abundant.

Feature coverage is a representation-level proxy: by itself, it does not guarantee improved downstream behavior. Its value comes from the hypothesis that post-training data are more effective when they activate safety-relevant directions that are missing or weakly supported in the original supervision. We therefore evaluate whether the coverage gains from feature-driven synthesis translate into improved safety behavior after SFT under a fixed data budget.

Taken together, these results show that SAE features are useful not only for analysis but also for data synthesis. They provide a concrete notion of representation-level coverage for prioritizing examples and enable a controllable synthesis pipeline. By improving coverage of safety-relevant internal directions, feature-driven synthesis yields a better safety--utility tradeoff after SFT and provides a useful coverage-based prioritization signal for future post-training tasks.
\clearpage

\section{Application: Supervised Fine-tuning}
\label{sec:sft}
\begin{figure}[h]
  \centering
  \includegraphics[width=1.0\linewidth]{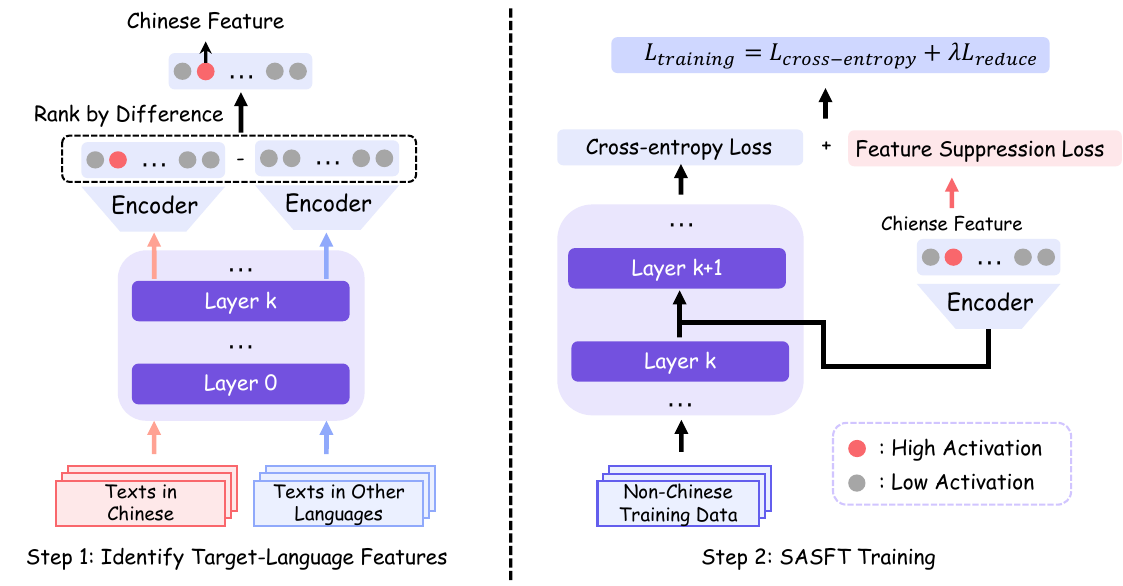}
    \caption{Overview of the $\textbf{S}$parse $\textbf{A}$utoencoder-guided $\textbf{S}$upervised $\textbf{F}$ine$\textbf{t}$uning (SASFT). SASFT operates in two steps: First, it identifies language-specific features in LLMs (left), then leverages these features as training signals to reduce code-switching behavior (right).}
    \label{fig:sasft}
\end{figure}
\begin{figure}[b] 
\centering
  \includegraphics[width=1.0\textwidth]{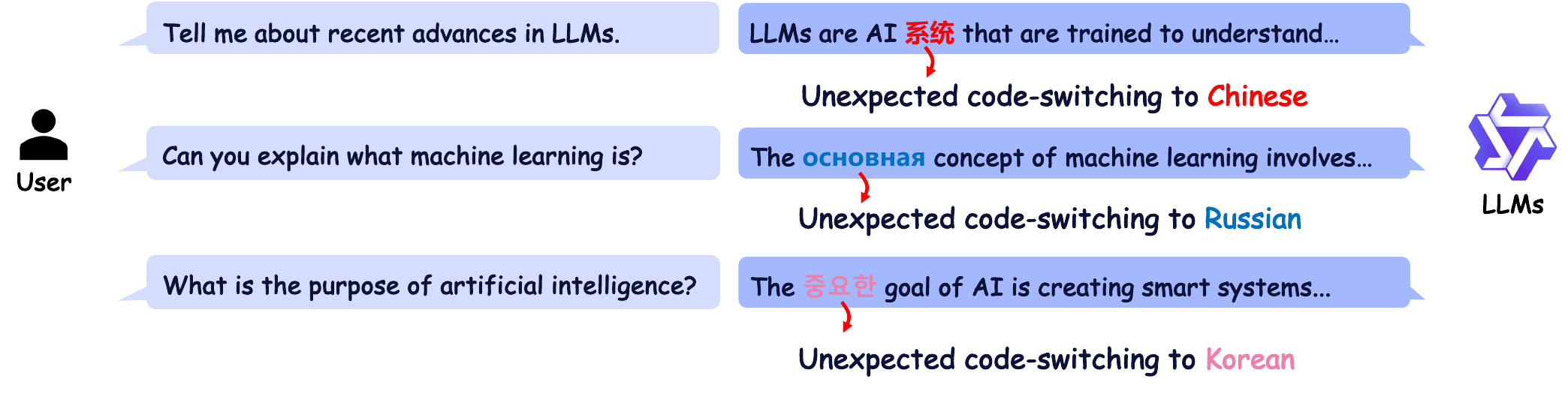}
  \caption{Examples of unexpected code-switching to Chinese, Russian, and Korean.}
  \label{fig:cs_example}
\end{figure}
Most existing works leverage SAEs for inference-time activation steering, which modifies the model's intermediate representations without updating its underlying parameters. Such test-time interventions offer no persistent improvement to the model itself and may compromise performance on unrelated tasks. This motivates us to explore whether SAEs can be leveraged to more fundamentally improve model behavior through training.


In this section, we investigate this question in the context of unexpected code-switching, a low-frequency but practically important failure mode in multilingual LLMs, where the model unexpectedly produces text in an unintended language, as shown in Figure~\ref{fig:cs_example}. Such failures are inherently challenging for standard SFT, because the supervision only encourages the model to match the target response and does not provide an explicit negative signal against undesired language switching. We find that SAEs provide an interpretable mechanism for identifying the language-specific internal features associated with this behavior. Based on this finding, we propose an SAE-guided supervised fine-tuning approach that reduces code-switching by explicitly suppressing the corresponding feature activations during training~\citep{deng2026sasft}.


\subsection{Unexpected Code-Switching}

Unexpected code-switching refers to the phenomenon where LLMs generate tokens in an unexpected language during response generation. Given a multilingual LLM $L$, an unexpected code-switching language $l$, and a set of prompts $\mathcal{X}=\{x_1,x_2,\ldots x_N\}$ where responses should not contain language $l$, we define code-switching ratio as follows:
\begin{align}
    r = \frac{1}{N}\sum_{i=1}^{N}\mathbb{I}(CSW(l,P_{L}(x_i))).\label{eq:cs_ratio}
\end{align}
Here, the function $CSW(l,y)$ checks if text $y$ contains any content in language $l$. $P_{L}(x_i)$ is the output when prompting $x_i$ to LLM $L$, and $\mathbb{I}(\cdot)$ denotes indicator function.

\subsection{Feature Analysis}
\begin{figure}[t] 
\centering
  \begin{subfigure}[t]{0.49\textwidth}
    \centering
    \includegraphics[width=\textwidth]{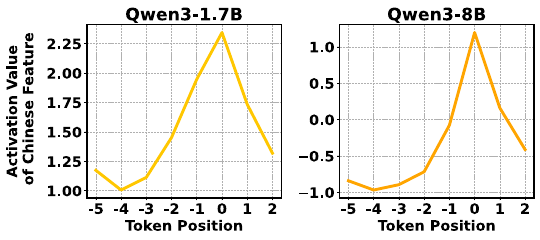}
    \vspace{0pt}
    \caption{The average pre-activation values of the Chinese feature at different token positions on responses with code-switching to Chinese. Position 0 represents the first token switching to Chinese.}
    \label{fig:zh_sae_diff_pos}
  \end{subfigure}
  \hfill
  \begin{subfigure}[t]{0.49\textwidth}
    \centering
    \includegraphics[width=\textwidth]{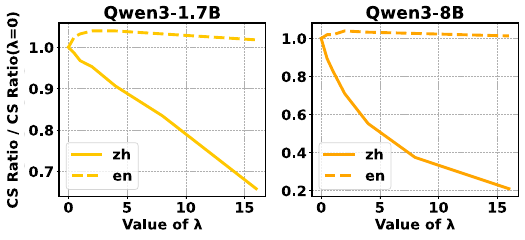}
    \vspace{0pt}
    \caption{The code-switch ratio to Chinese after ablating Chinese/English features with different ablation coefficient $\lambda$. }
    \label{fig:reduce_sae}
  \end{subfigure}
  \caption{Analysis of the language feature and its role in code-switching.}
  \label{fig:combined}
\end{figure}

Building on the language-specific features identified via SAEs~\citep{deng2025unveilinglanguagespecificfeatureslarge}, we conduct a mechanistic analysis of unexpected code-switching. Two key findings motivate our method.

\textbf{Pre-activation values rise before code-switching.} We take code-switching to Chinese as a representative case and track the average pre-activation value of the Chinese language feature at each token position relative to the first code-switched token (position 0). As shown in Figure~\ref{fig:zh_sae_diff_pos}, the pre-activation values gradually increase in the tokens leading up to position 0, and peak at the switch, consistently across all five models. This suggests that abnormally high pre-activation values may serve as a precursor to unexpected code-switching.

\textbf{Directional ablation of language features suppresses code-switching.} We apply directional ablation~\citep{DBLP:conf/iclr/FerrandoORN25,DBLP:conf/nips/ArditiOSPPGN24} to subtract the target language feature direction from the residual stream $\mathbf{x} \in \mathbb{R}^N$ at the final layer of the token immediately preceding the first unexpected code-switching token. This process can be expressed as:
\begin{equation}
    \mathbf{x}'\leftarrow\mathbf{x}-\lambda \mathbf{d},\label{eq:ablation}
\end{equation}
where $\mathbf{d}$ represents the language feature and $\lambda$ is the coefficient that controls the degree of ablation. After obtaining $\mathbf{x}'$, we replace $\mathbf{x}$ with $\mathbf{x}'$ and continue the forward pass of the LLMs. As shown in Figure~\ref{fig:reduce_sae}, it consistently reduces the code-switching ratio, with larger ablation coefficients yielding greater reductions. In contrast, ablating an irrelevant language feature has a negligible effect, confirming the language-specificity of the identified features. 

\subsection{Method}

While inference-time ablation demonstrates that suppressing language-specific feature activations can mitigate code-switching, it requires external intervention at every decoding step and fails to address the root cause within the model parameters. To overcome these limitations, we propose \textbf{Sparse Autoencoder-guided Supervised Fine-Tuning (SASFT)}, which internalizes feature suppression directly into the training process.

SASFT operates in two stages. First, language-specific features for a target language $L$ are identified by ranking SAE features according to a monolinguality score $\nu_s^L = \mu_s^L - \gamma_s^L$, where $\mu_s^L$ and $\gamma_s^L$ denote the mean activation of feature $s$ on language-$L$ data and all other languages, respectively. Second, an auxiliary regularization loss is introduced alongside the standard cross-entropy objective. Formally, consider a language $L$ that we aim to avoid code-switching to. We have sets of residual streams $\mathcal{D}=\{\mathcal{D}_1, \ldots, \mathcal{D}_K\}$, where each $\mathcal{D}_i$ contains the residual streams from training data in language $i$ for a specific layer. The auxiliary loss can be defined as follows:
\begin{align}
    L_{\text{reduce}} = \mathbb{E}_{\mathcal{D}_j \sim \mathcal{D} \setminus \{\mathcal{D}_L\}} \left[ \mathbb{E}_{\mathbf{x} \sim \mathcal{D}_j} \left[ \sum\limits_{s \in \mathcal{S}_L}\mathrm{ReLU} \left( \mathbf{f}_s(\mathbf{x}) - \alpha_j \right)\right] \right] ,\label{eq:aux_loss_reduce}
\end{align}
where $\mathbf{f}_s(\mathbf{x})$ is the pre-activation values of feature $s$ for the residual stream $\mathbf{x}$. The set $\mathcal{S}_L$ denotes the language-specific features for language $L$. For each feature $s$ in language $j$, we use $\alpha_j$ to represent its pre-estimated average pre-activation value. We don't set $\alpha_j$ to zero because the pre-estimated average pre-activation value can be negative. In such cases, zero would be too large as a baseline value. Additionally, $\mathcal{D}_L$ is the set of residual streams for language $L$, which we exclude because generating language $L$ from language $L$ does not count as code-switching.

For SASFT, we combine two losses to get the final training loss:
\begin{align}
    L_{training} = L_{\text{cross-entropy}} + \lambda L_{\text{reduce}}
\end{align}
where $\lambda$ is a hyperparameter we can adjust to control how much $L_{\text{reduce}}$ contributes to the total loss.

\subsection{Main Results}
We evaluate SASFT on five models spanning three model families (Gemma-2, Llama-3.1, and Qwen3) across three target languages (Chinese, Russian, and Korean). As shown in Table~\ref{tab:cs_ratio_main}, SASFT consistently outperforms all baselines across both dataset settings, achieving over 50\% reduction in code-switching ratio in the majority of experimental settings, with complete elimination in certain configurations (e.g., Qwen3-1.7B on Korean). Table~\ref{tab:overall_performance} further shows that SASFT maintains or marginally improves performance across six multilingual benchmarks, confirming that suppressing undesirable language features does not compromise general multilingual competence.

\begin{table*}[t]
\caption{Comparison of code-switching ratios (\%) across different methods and models. For each target language (Chinese, Russian, and Korean), we train models on two dataset settings: a 210k dataset and a 110k dataset, then evaluate their code-switching ratio to Chinese, Russian, and Korean. \textbf{Bold} numbers indicate the best results. Results show SASFT consistently outperforms the baselines, achieving over 50\% reduction in most cases. 
\label{tab:cs_ratio_main}\\}
\centering
\setlength{\tabcolsep}{2mm}{
\begin{adjustbox}{max width=\textwidth}
\begin{tabular}{llllllll}
\toprule
\multirow{2}{*}{Model} & \multirow{2}{*}{Method} &\multicolumn{3}{c}{Training Data 210k} & \multicolumn{3}{c}{Training Data 110k} \\
\cmidrule(lr){3-5} \cmidrule(lr){6-8}
&& CS: any $\to$ zh & CS: any $\to$ ru& CS: any $\to$ ko & CS: any $\to$ zh& CS: any $\to$ ru & CS: any $\to$ ko\\
\midrule
\multirow{4}{*}{Qwen3-1.7B-Base} & SFT (Baseline)  &0.81&0.19&0.36&0.68&0.19&0.23\\
& SFT+GRPO &0.66\ \ (-19\%)&0.11\ \ (-42\%)&0.34\ \ (-6\%)&0.68\ \ (0\%)&0.19\ \ (+1\%)&0.20\ \ (-16\%)\\
& SFT+Penalty &0.53\ \ (-35\%)&0.09\ \ (-53\%)&0.06\ \ (-84\%)&0.49\ \ (-28\%)&0.07\ \ (-62\%)&0.06\ \ (-73\%)\\
& SASFT &\textbf{0.22\ \ (-72\%)}&\textbf{0.03\ \ (-85\%)}&\textbf{0.00\ \ (-100\%)}&\textbf{0.31\ \ (-55\%)}&\textbf{0.03\ \ (-87\%)}&\textbf{0.02\ \ (-93\%)}\\
\cmidrule(lr){1-8} 
\multirow{4}{*}{Qwen3-8B-Base} & SFT (Baseline)  &0.96&0.16&0.43&0.83&0.17&0.25\\
& SFT+GRPO &0.70\ \ (-14\%)&0.09\ \ (-40\%)&0.22\ \ (-27\%)&0.67\ \ (-26\%)&0.06\ \ (-65\%)&0.12\ \ (-20\%)\\
& SFT+Penalty &0.70\ \ (-27\%)&0.12\ \ (-24\%)&0.23\ \ (-47\%)&0.76\ \ (-9\%)&0.08\ \ (-50\%)&0.18\ \ (-27\%)\\
& SASFT &\textbf{0.66\ \ (-31\%)}&\textbf{0.07\ \ (-56\%)}&\textbf{0.07\ \ (-83\%)}&\textbf{0.62\ \ (-26\%)}&\textbf{0.07\ \ (-59\%)}&\textbf{0.05\ \ (-80\%)}\\

\bottomrule
\end{tabular}
\end{adjustbox}
}
\end{table*}

\begin{table*}[t]
\caption{Performance comparison on six benchmarks across different methods. We evaluate models trained on the Chinese 110k dataset setting. Results demonstrate that SASFT successfully maintains model capabilities while reducing code-switching, even showing improvements in several cases. The \textcolor{red}{red numbers} indicate performance improvements compared to the SFT.
\label{tab:overall_performance}\\}
\centering
\setlength{\tabcolsep}{1.5mm}{
\begin{adjustbox}{max width=\textwidth}
\begin{tabular}{lllllllll}
\toprule

 \multirow{2}{*}{Model} & \multirow{2}{*}{Method} & MMLU & HumanEval & Flores & HellaSwag & LogiQA & IFEval & MGSM\\
\cmidrule(lr){3-9} 
&& Acc (\%)& Acc (\%)& Bleu (\%)& Acc (\%)& Acc (\%)& Acc (\%)& Acc (\%)\\
\midrule
\multirow{4}{*}{Qwen3-1.7B-Base} 
& SFT &37.47&90.29&23.70&33.53&32.38&20.27&32.91\\
& SFT+GRPO &37.80 (\textcolor{red}{+0.33})&90.48 (\textcolor{red}{+0.19})&23.45 (-0.25)&35.74 (\textcolor{red}{+2.21})&31.37 (-1.01)&20.19 (-0.08)&32.67 (-0.24)\\
& SFT+Penalty & 37.78 (\textcolor{red}{+0.31}) & 89.13 (-1.16) & 23.55 (-0.15) & 36.24 (\textcolor{red}{+2.71}) & 33.00 (\textcolor{red}{+0.62}) & 20.44 (\textcolor{red}{+0.17}) & 33.60 (\textcolor{red}{+0.69}) \\
& SASFT &38.38 (\textcolor{red}{+0.91})&89.04 (-1.25)&23.67 (-0.03)&33.71 (\textcolor{red}{+0.18})&32.38 (0.00)&20.22 (-0.05)&30.85 (-2.06)\\
\cmidrule(lr){1-9} 
\multirow{4}{*}{Qwen3-8B-Base} 
& SFT &52.15&95.87&29.99&42.48&42.25&33.64&58.03\\
& SFT+GRPO &50.85 (-1.30)&96.44 (\textcolor{red}{+0.57})&30.14 (\textcolor{red}{+0.15})&44.48 (\textcolor{red}{+2.00})&41.50 (-0.75)&33.42 (-0.22)&55.28 (-2.75)\\
& SFT+Penalty & 50.74 (-1.41) & 94.71 (-1.16) & 30.10 (\textcolor{red}{+0.11}) & 34.51 (-7.97) & 39.88 (-2.37) & 34.04 (\textcolor{red}{+0.40}) & 56.29 (-1.74) \\
& SASFT &50.09 (-2.06)&98.27 (\textcolor{red}{+2.40})&29.97 (-0.02)&39.60 (-2.88)&42.75 (\textcolor{red}{+0.50})&33.91 (\textcolor{red}{+0.27})&58.45 (\textcolor{red}{+0.42})\\

\bottomrule
\end{tabular}
\end{adjustbox}
}

\vspace{-4pt}
\end{table*}

\clearpage

\section{Application: Reinforcement Learning}
\label{sec:rl}
\begin{figure}[h]
  \centering
  \includegraphics[width=1.0\linewidth]{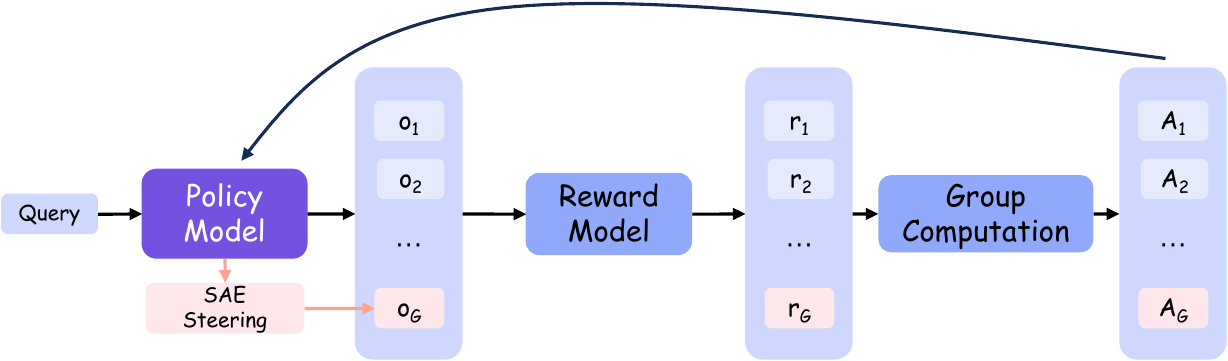}
    \caption{Overview of SAE-guided DAPO with rare-negative augmentation. The policy model generates $G-1$ normal outputs and one additional output steered by SAE feature intervention to serve as a rare negative sample.
}
    \label{fig:dapo}
\end{figure}
Beyond SFT, we also explore integrating SAEs into the online RL pipeline. Before presenting our approach, we briefly describe our initial attempts and the lessons learned.

\textbf{Early Attempt}: SAE-Guided Positive Rollout Generation. We initially attempted to use SAE feature steering to generate higher-quality positive rollouts. However, this direction proved challenging: steering alone is insufficient to produce correct responses for tasks requiring precise multi-step reasoning, and it may compromise the fluency of generated text, potentially causing the model to learn from unnatural patterns and thereby degrading general performance.

\textbf{Revised Approach}: SAE-Guided Rare Negative Augmentation. These challenges motivated us to shift focus toward negative sample generation. SAE steering is particularly well-suited for this purpose: undesirable behaviors are easier to induce than correct ones, and any fluency degradation is inconsequential since the model learns to avoid rather than imitate these samples. 

We therefore focus on endless repetition as a representative low-frequency failure mode, and use SAE feature steering to augment the rollout distribution with rare negative samples, providing explicit training signal against behaviors that are otherwise difficult to correct. This augmentation is crucial because standard online RL rarely encounters such failure cases during rollouts due to their low occurrence probability, and therefore provides only weak signal for eliminating them.

\subsection{Feature Analysis}
\label{sec:feature_analysis}
Endless repetition is characterized by a self-reinforcing pattern, where the model becomes increasingly trapped in a loop of repeated content. We therefore hypothesize that certain SAE features are specifically associated with this process, and that their activation values are progressively amplified as repetition continues. To validate this hypothesis, we conduct the following experiments.

\begin{figure}[ht] 
\centering
  \includegraphics[width=1.0\textwidth]{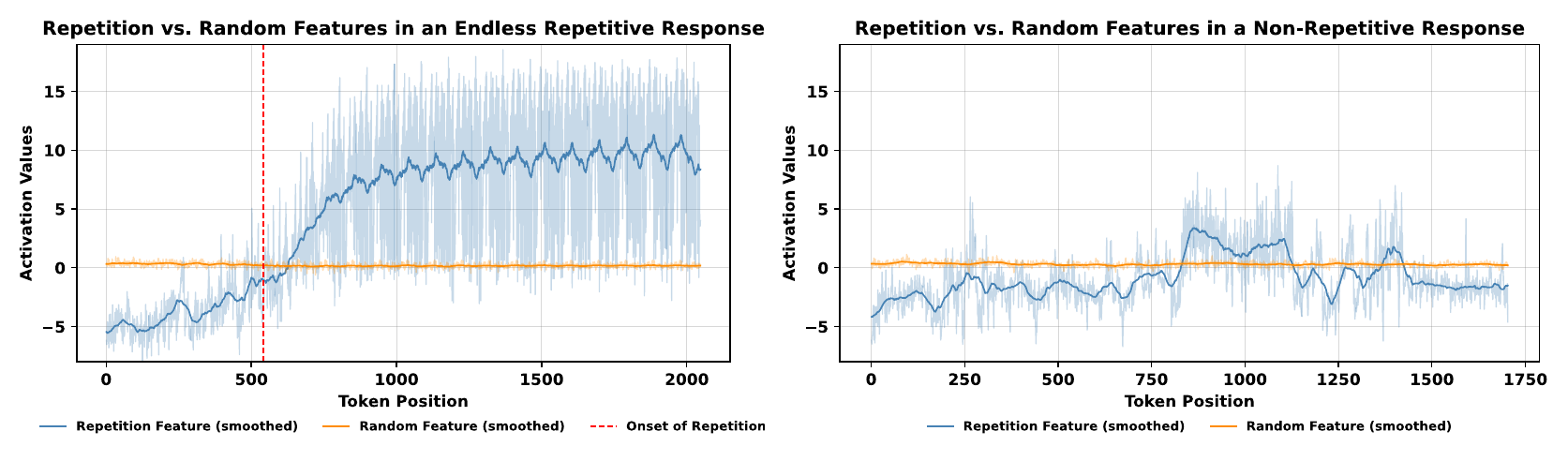}
  \caption{Activation values of a repetition feature and a randomly selected feature over token positions in a repetitive response (left) and a non-repetitive response (right). In the repetitive response, the repetition feature exhibits a sharp and sustained increase around the onset of repetition (red dashed line), while remaining near zero in the non-repetitive response, consistent with the random feature in both cases. (Model: Qwen3-8B)}
  \label{fig:repeat_feature}
\end{figure}

\textbf{Identifying Repetition Features.} We collect samples where the model spontaneously generates endless repetitive content. For each repeated token, we compute the difference in SAE feature activations between its first occurrence and its last repeated occurrence within a given context. The rationale for comparing the same token is that it controls for token-specific variations, ensuring that the observed activation differences are more likely attributable to the repetition process itself rather than to differences in token identity. Features with the largest activation increases are identified as repetition-related features. As shown in Figure~\ref{fig:repeat_feature}, certain features (repetition features) exhibit a sharp increase in activation values and remain persistently elevated during endless repetition, whereas in non-repetitive responses they stay near zero throughout.

\begin{figure}[ht] 
\centering
  \includegraphics[width=0.8\textwidth]{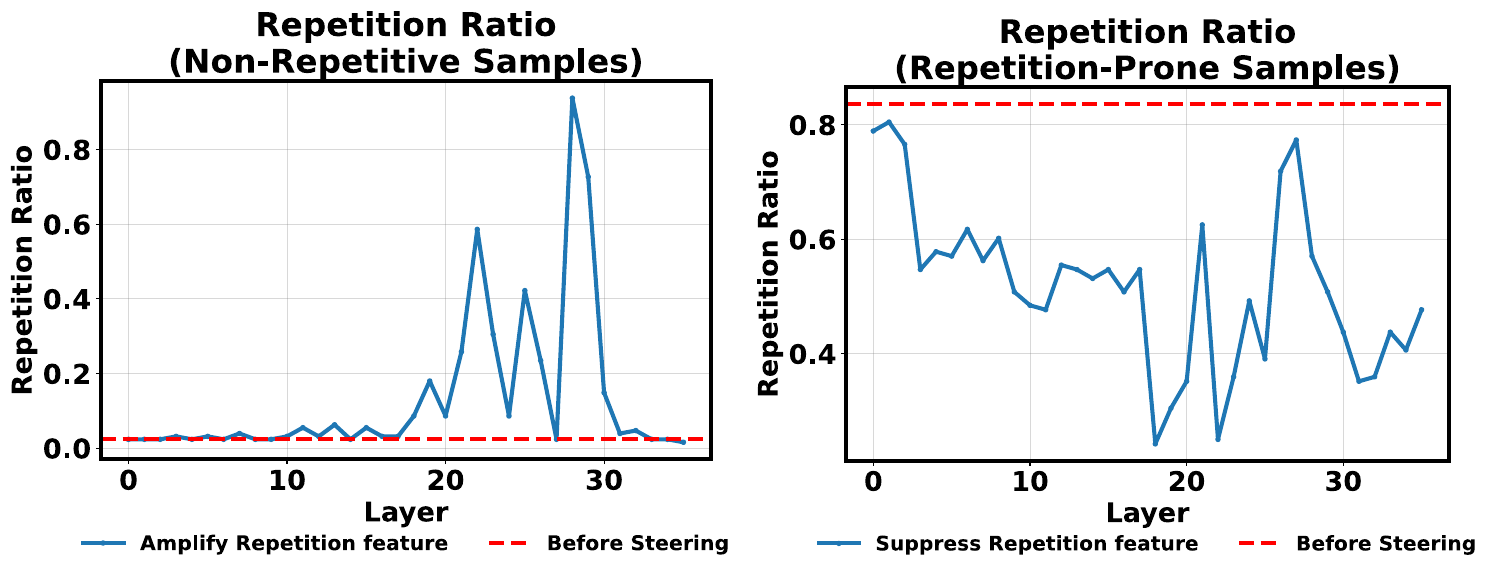}
  \caption{SAE feature steering controls repetition ratio across layers. Amplifying the repetition feature on non-repetitive samples increases repetition (left), while suppressing it on repetition-prone samples reduces repetition below the baseline (right), confirming the causal role of the features. (Model: Qwen3-8B)}
  \label{fig:repeat_steering}
\end{figure}

\textbf{Causal Verification via Steering.} To establish a causal relationship between the identified features and repetitive behavior, we conduct bidirectional steering experiments. As shown in Figure~\ref{fig:repeat_steering}, suppressing these features on repetitive samples leads to a consistent reduction in repetition rate, while amplifying them on normal samples successfully induces repetitive behavior. These results confirm that the identified features are causally linked to endless repetition rather than merely correlated with it.

\begin{figure}[ht] 
\centering
  \includegraphics[width=1.0\textwidth]{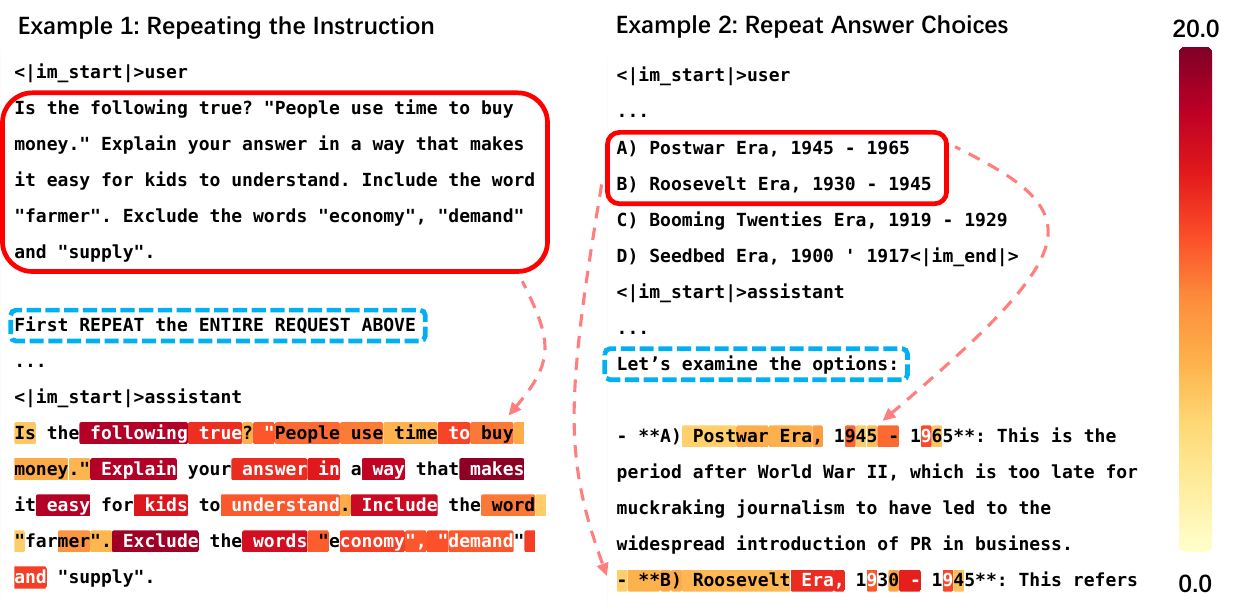}
  \caption{SAE feature activation heatmap on Qwen3-8B for two benign repetition scenarios (tokens with activation $>$ 5.0 are highlighted). \textbf{Example 1}: The model repeats the user's instruction as requested. \textbf{Example 2}: The model reproduces answer choices in a multiple-choice task. The endless repetition features show high activation in both cases, suggesting that the identified repetition features may also be associated with normal repetitive behavior.}
  \label{fig:repeat_heat}
\end{figure}

\textbf{Feature Semantics: Beyond Endless Repetition.} We initially assumed that endless repetition and benign repetition would be governed by distinct features, as they represent fundamentally different phenomena: the former is a form of model output collapse, while the latter is a normal and expected behavior. However, as illustrated in Figure~\ref{fig:repeat_heat}, we find that the same features exhibit high activation values in benign repetition scenarios as well, such as when the model is asked to repeat a user's question, or when it reproduces answer choices in multiple-choice tasks. This suggests that the identified features capture a more general notion of repetition rather than being exclusive to pathological cases. This is also why we do not adopt the approach described in Section~\ref{sec:sft} to address endless repetition: since the repetition features are shared between endless and benign repetition, directly suppressing their activations during training would risk degrading the model's ability to perform normal repetitive behavior.
\subsection{Method}
We build our approach on top of DAPO~\citep{yu2025dapo} without Dynamic Sampling\footnote{We disable Dynamic Sampling because it can make the time cost of each training step longer and less controllable. Hereafter, we use DAPO to denote DAPO without Dynamic Sampling.}. The core idea is to augment the rollout distribution with synthetic negative samples by leveraging SAE feature steering to induce repetitive behavior.

\textbf{SAE Feature Steering.} In Section~\ref{sec:feature_analysis}, we identify SAE features that are causally linked to endless repetition. Here, we leverage these features to steer the model toward generating repetitive content. Specifically, we use \textit{feature steering} to add the repetition feature to the residual stream $\mathbf{h} \in \mathbb{R}^N$ at each generation step. This process can be expressed as:
\begin{equation}
    \mathbf{h}' \leftarrow \mathbf{h} + \alpha \mathbf{d},
    \label{eq:steering}
\end{equation}
where $\mathbf{d}$ represents the repetition feature direction and $\alpha$ is the steering coefficient that controls the degree of amplification. After obtaining $\mathbf{h}'$, we replace $\mathbf{h}$ with $\mathbf{h}'$ and continue the forward pass of the model. A larger $\alpha$ leads to stronger repetitive behavior in the generated output.

\textbf{Rollout Augmentation.} Concretely, for each group of rollouts, we sample $G-1$ outputs normally from the policy model, and apply SAE feature steering with coefficient $\alpha$ to generate one additional output $o_G$, which is expected to exhibit repetitive behavior. This steered rollout is then incorporated into the group alongside the normal rollouts, providing an explicit training signal against endless repetition that would otherwise be rarely encountered during standard RL training. The full procedure is summarized in Algorithm~\ref{algo:sae_dapo}.

\begin{table}[h]
    \centering
    \begin{tabular}{@{}p{1.0\textwidth}@{}} 
        \toprule 
        \textbf{Algorithm:}~~~\textbf{SAE-Guided DAPO with Rare Negative Augmentation} \\
        \midrule 
        \textbf{Input} initial policy model $\pi_\theta$; reward model $R$; task prompts $\mathcal{D}$; \textcolor{red}{SAE feature set $\mathcal{F}$; steering coefficient $\alpha$}; hyperparameters $\varepsilon_\mathtt{low}, \varepsilon_\mathtt{high}$ \\
        ~1: \textbf{for} step = 1,...,M \textbf{do} \\
        ~2: ~~~ Sample a batch $\mathcal{D}_b$ from $\mathcal{D}$ \\
        ~3: ~~~ Update the old policy model $\pi_{\theta_{old}} \leftarrow \pi_\theta$\\
        \textcolor{red}{~4: ~~~ \textbf{for} each question $q \in \mathcal{D}_b$ \textbf{do}} \\
        \textcolor{red}{~5: ~~~~~~~ Sample $G-1$ outputs $\{o_i\}_{i=1}^{G-1} \sim \pi_{\theta_{\text{old}}}(\cdot | q)$ normally} \\
        \textcolor{red}{~6: ~~~~~~~ Sample one additional output $o_G$ with SAE feature steering on $\mathcal{F}$ with coefficient $\alpha$} \\
        \textcolor{red}{~7: ~~~~~~~ Set $\{o_i\}_{i=1}^{G} = \{o_1, ..., o_{G-1}, o_G\}$} \\
        \textcolor{red}{~8: ~~~ \textbf{end for}} \\
        ~9: ~~~ Compute rewards $\{r_i\}_{i=1}^{G}$ for each sampled output $o_i$ by running $R$ \\
        10: ~~ For each sampled output $o_i$, compute $\hat{A}_{i,t}$ for the \textit{t}-th token of $o_i$ \\
        11: ~~ \textbf{for} iteration = 1, ..., $\mu$ \textbf{do}\\
        12: ~~~~~~~ Update the policy model $\pi_\theta$ by maximizing the DAPO objective \\
        \textbf{Output} $\pi_\theta$\\
        \bottomrule
    \end{tabular}
    \captionsetup{labelformat=empty}
    \caption{}
    \label{algo:sae_dapo}
\end{table}

\subsection{Experimental Setting}

We evaluate SAE-guided rare negative augmentation in the online RL stage on three models of different scales: Qwen3-1.7B, Qwen3-8B, and Qwen3-30B-A3. For all three models, the RL starting point (\textit{Before RL}) is a cold-start model obtained by supervised fine-tuning on a set of SFT data. We compare our method against vanilla DAPO under the same RL setup, with the only difference being that our method augments each rollout group with one SAE-steered negative sample that is biased toward repetitive behavior.

To measure the target failure mode, we track the \textit{repeat ratio} during RL training, defined as the fraction of sampled responses that exhibit endless repetition when generating on a held-out set of roughly 10,000 prompts. In addition, to assess whether the intervention affects broader model capability, we evaluate the post-RL models on a suite of standard benchmarks, including MMLU, Flores, HellaSwag, LogiQA, IFEval, and MGSM.

\subsection{Main Results}

\begin{figure}[t] 
\centering
  \includegraphics[width=1.0\textwidth]{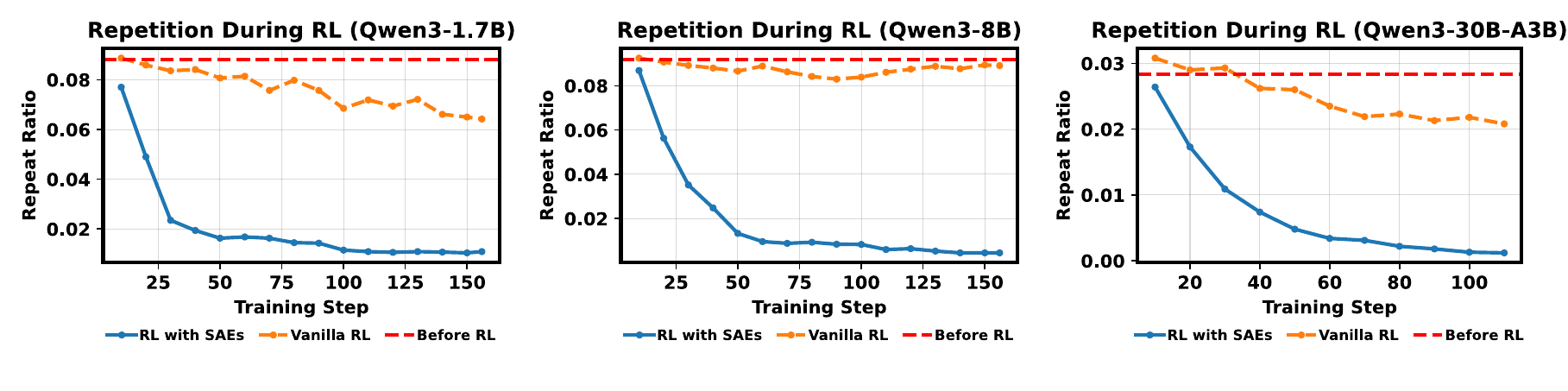}
  \caption{Repetition ratio during RL training for Qwen3-1.7B, Qwen3-8B, and Qwen3-30B-A3. Compared with vanilla RL, SAE-guided rare negative augmentation (\textit{RL with SAEs}) consistently reduces repetition much faster and to a substantially lower level across all model sizes. The red dashed line indicates the repetition ratio before RL. These results show that explicitly injecting SAE-steered repetitive rollouts provides an effective training signal against this otherwise under-represented failure mode.}
  \label{fig:rl_main}
\end{figure}

Figure~\ref{fig:rl_main} shows that SAE-guided rare negative augmentation consistently reduces repetition much more effectively than vanilla RL across all three model scales. In all cases, the repeat ratio under our method drops sharply in the early stage of training and continues to decrease to a very low level. By contrast, vanilla RL yields only limited improvement: although it sometimes reduces repetition slightly relative to the pre-RL model, the overall decrease remains modest, and the repeat ratio stays substantially higher than that achieved by our method throughout training. These results support our central motivation that endless repetition is a low-frequency failure mode that is insufficiently represented in standard rollout distributions, making it difficult for vanilla RL to learn a strong corrective signal. By explicitly injecting SAE-steered repetitive rollouts, our method increases the visibility of this failure mode during training and enables the policy to learn to avoid it more effectively.

Table~\ref{tab:qwen_rl_sae_performance} reports the downstream benchmark results after RL. Overall, SAE-guided RL remains broadly competitive with vanilla RL on general capability benchmarks, while providing a much stronger reduction in repetition. At the same time, the effect on downstream performance is mixed and task-dependent: some benchmarks show small gains relative to vanilla RL or the pre-RL model, while others exhibit regressions. Taken together, these results suggest that SAE-guided rare negative augmentation is effective at targeting the intended failure mode during RL, but does not uniformly improve general-purpose capability. Its main benefit lies in supplying an explicit negative training signal for a rare pathological behavior that standard RL alone does not adequately cover.

\begin{table*}[t]
\caption{Main evaluation results after RL on three Qwen3 models. We compare the base model before RL, vanilla RL, and our SAE-guided RL method. Numbers in parentheses denote the change relative to the \textit{Before RL} baseline. Overall, SAE-guided rare negative augmentation preserves competitive general capabilities while providing an explicit training signal against repetition. The \textcolor{red}{red numbers} indicate performance improvements compared to Before RL.}
\vspace{-4pt}
\centering
\setlength{\tabcolsep}{1.5mm}{
\begin{adjustbox}{max width=\textwidth}
\begin{tabular}{llllllll}
\toprule
\multirow{2}{*}{Model} & \multirow{2}{*}{Method} & MMLU & Flores & HellaSwag & LogiQA & IFEval & MGSM \\
\cmidrule(lr){3-8}
&& Acc (\%) & Bleu (\%) & Acc (\%) & Acc (\%) & Acc (\%) & Acc (\%) \\
\midrule

\multirow{3}{*}{Qwen3-1.7B}
& Before RL & 41.78 & 28.47 & 39.66 & 34.62 & 42.29 & 46.80 \\
& Vanilla RL & 41.83 (\textcolor{red}{+0.05}) & 29.44 (\textcolor{red}{+0.97}) & 39.99 (\textcolor{red}{+0.32}) & 36.12 (\textcolor{red}{+1.50}) & 40.10 (-2.19) & 46.48 (-0.32) \\
& RL+SAE & 41.67 (-0.10) & 31.06 (\textcolor{red}{+2.59}) & 40.93 (\textcolor{red}{+1.26}) & 34.88 (\textcolor{red}{+0.25}) & 40.42 (-1.88) & 52.36 (\textcolor{red}{+5.56}) \\
\cmidrule(lr){1-8}

\multirow{3}{*}{Qwen3-8B}
& Before RL & 48.40 & 38.18 & 61.02 & 48.00 & 71.04 & 70.12 \\
& Vanilla RL & 48.55 (\textcolor{red}{+0.15}) & 38.32 (\textcolor{red}{+0.13}) & 60.85 (-0.17) & 47.12 (-0.88) & 70.42 (-0.62) & 70.96 (\textcolor{red}{+0.84}) \\
& RL+SAE & 48.40 (0.00) & 38.74 (\textcolor{red}{+0.55}) & 61.97 (\textcolor{red}{+0.95}) & 47.12 (-0.88) & 68.96 (-2.08) & 72.40 (\textcolor{red}{+2.28}) \\
\cmidrule(lr){1-8}

\multirow{3}{*}{Qwen3-30B-A3B}
& Before RL & 51.75 & 40.91 & 70.16 & 48.12 & 71.98 & 76.56 \\
& Vanilla RL & 51.43 (-0.33) & 40.87 (-0.04) & 69.53 (-0.63) & 48.38 (\textcolor{red}{+0.25}) & 72.29 (\textcolor{red}{+0.31}) & 77.64 (\textcolor{red}{+1.08}) \\
& RL+SAE & 52.23 (\textcolor{red}{+0.48}) & 40.84 (-0.06) & 69.21 (-0.95) & 48.75 (\textcolor{red}{+0.63}) & 71.25 (-0.73) & 82.40 (\textcolor{red}{+5.84}) \\
\bottomrule
\end{tabular}
\end{adjustbox}
}

\label{tab:qwen_rl_sae_performance}
\end{table*}

\clearpage

\section{Conclusion}
\label{sec:conclusion}
\subsection{Summary}
In this report, we introduced \textbf{Qwen-Scope}, an open-source suite of sparse autoencoders for the Qwen model family. Qwen-Scope provides layer-wise SAE features for multiple Qwen3 and Qwen3.5 backbones, covering both dense and mixture-of-experts architectures under a unified training pipeline.

We demonstrated that Qwen-Scope is useful not only for post-hoc interpretation, but also for practical model-development workflows. By releasing these modules and concrete use cases, we hope to support community-driven exploration of Qwen-series models and enable researchers and developers to uncover new mechanisms and applications beyond those presented in this report.

\subsection{Exploring Directions}

Qwen-Scope opens several directions for future research. We highlight a few directions that are especially valuable for connecting interpretability tools to more controllable, and more useful model development.

\paragraph{Reasoning-model interpretability.}
As models increasingly rely on long chain-of-thought reasoning, multi-step sampling, and potentially latent or vector-space reasoning, analyzing a single forward pass may be insufficient. Qwen-Scope can help study which SAE features appear across reasoning branches, which steps are causally important, and how internal reasoning trajectories change under resampling or intervention~\citep{macar2026thoughtbranchesinterpretingllm,bogdan2025thoughtanchorsllmreasoning}.

\paragraph{Internals-based monitoring and auditing.}
SAE features may provide lightweight internal signals for risks that are difficult to detect from outputs alone, such as deception, hidden objectives, jailbreak susceptibility, and hallucination. Future work can combine Qwen-Scope with probes, activation-based monitors, and auditing pipelines to test whether internal representations reveal such risks early and robustly~\citep{goldowskydill2025detectingstrategicdeceptionusing,parrack2026benchmarkingdeceptionprobesblacktowhite,marks2025auditinglanguagemodelshidden}.

\paragraph{Model diffing and post-training analysis.}
Qwen-Scope can be used to compare model internals before and after fine-tuning, reinforcement learning, or other interventions. Instead of only measuring behavioral changes, researchers can analyze which SAE features change, which directions become more or less active, and whether post-training leaves readable traces in activation space~\citep{minder2026narrow}.

\paragraph{Interpretability-driven control and training.}
The results in this report suggest that SAE features can act as control knobs: they can be amplified or suppressed at inference time, used as auxiliary signals during SFT, or used to construct rare negative examples for RL. Future work can further study how feature-level interventions affect generalization, robustness, and safety, and how interpretable directions can be incorporated into training pipelines~\citep{casademunt2025steeringoutofdistributiongeneralizationconcept}.

\paragraph{Data-centric interpretability.}
Qwen-Scope can also support data-centric workflows by connecting training data to internal feature coverage. Future work can use SAE features to identify under-covered behaviors, prioritize examples, guide synthetic data generation, and attribute undesirable behavior to influential data regions~\citep{coalson2025ifguideinfluencefunctionguideddetoxification,li2024influencefunctionsworklarge}.

We welcome the community to use Qwen-Scope to explore these and other application directions. We hope that open SAE for Qwen-series models will make it easier to study model internals, unexpected behaviors, and build new workflows that connect interpretability research to practical model improvement.

\subsection{Social Impact}
We acknowledge that current interpretability research does not yet provide sufficient safeguards against misuse. We strongly urge developers and researchers to refrain from applying Qwen-Scope or Qwen models in any manner that violates human ethical values.
It is strictly prohibited to use interpretability tools for non-scientific research purposes to interfere with model capabilities, or to fabricate, generate, and disseminate harmful information that violates public order, good morals, and socialist core values, including pornographic, violent, discriminatory, or incendiary content. Violators will have their authorization automatically terminated and shall bear all legal liabilities arising therefrom. The right of final interpretation of this statement belongs to the project owner.
\clearpage

\section*{Authors}
Core contributors~\footnote{Boyi Deng, Xu Wang, and Yaoning Wang are in charge of experimental designs and empirical evidence. Yu Wan is the project leader. Yubo Ma and Baosong Yang are co-supervisors. We truly thank all team members for their insightful comments.}: Boyi Deng, Xu Wang, Yaoning Wang, Yu Wan, Yubo Ma, Baosong Yang. 

Contributors: Haoran Wei, Jialong Tang, Huan Lin, Ruize Gao, Tianhao Li, Qian Cao, Xuancheng Ren, Xiaodong Deng, An Yang, Fei Huang, Dayiheng Liu, Jingren Zhou.

\bibliography{biblio}
\bibliographystyle{colm2024_conference}

\end{document}